\documentclass[conference]{IEEEtran}
\IEEEoverridecommandlockouts
\usepackage{cite}
\usepackage{amsmath,amssymb,amsfonts}
\usepackage{algorithmic}
\usepackage{graphicx}
\usepackage{textcomp}
\usepackage{xcolor}
\usepackage{float}
\usepackage{booktabs}
\usepackage{subfigure}
\usepackage{amsmath}
\usepackage{epstopdf}
\usepackage{epsfig}
\usepackage{multirow}
\usepackage{hyperref}
\usepackage{xcolor}
\usepackage{colortbl} 
\usepackage{hhline} 
\usepackage{longtable}
\usepackage{lscape}
\usepackage{float}
\usepackage{algorithmic}
\usepackage{algorithm}
\usepackage{tabularx}
\hypersetup{hidelinks}

\def\BibTeX{{\rm B\kern-.05em{\sc i\kern-.025em b}\kern-.08em
		T\kern-.1667em\lower.7ex\hbox{E}\kern-.125emX}}
\begin{document}
	
\title{LLM-Enabled Low-Altitude UAV Natural Language Navigation via Signal Temporal Logic Specification Translation and Repair}

\author{Yuqi Ping, Huahao Ding, Tianhao Liang, Longyu Zhou, Guangyu Lei, Xinglin Chen,\\ Junwei Wu, Jieyu Zhou, Tingting Zhang
	
	\thanks{Yuqi Ping, Huahao Ding, Tianhao Liang, Guangyu Lei, Xinglin Chen, Junwei Wu, and T. Zhang are with Guangdong Provincial Key Laboratory of Space-Aerial Networking and Intelligent Sensing, Harbin Institute of Technology, Shenzhen, China, (e-mail: pingyq@stu.hit.edu.cn; hitszdhh@163.com; liangth@hit.edu.cn; GuangyuLei@stu.hit.edu.cn; chenxinglin@stu.hit.edu.cn; 220210419@stu.hit.edu.cn; zhangtt@hit.edu.cn); T. Zhang is also with Peng Cheng Laboratory (PCL), Shenzhen, China. Longyu Zhou is with the Information Systems Technology and Design, Singapore University of Technology and Design, Singapore 487372, (e-mail: zhoulyfuture@outlook.com). Jieyu Zhou is with School of Computer Science and Engineering, Central South University, Changsha, China, (e-mail: zhoujieyu@csu.edu.cn).}

}

\markboth{IEEE Transactions on Cognitive Communications and Networking}
{Shell \MakeLowercase{\textit{et al.}}: A Sample Article Using IEEEtran.cls for IEEE Journals}

\maketitle

\begin{abstract}
	Natural language (NL) navigation for low-altitude unmanned aerial vehicles (UAVs) offers an intelligent and convenient solution for low-altitude aerial services by enabling an intuitive interface for non-expert operators. However, deploying this capability in urban environments necessitates the precise grounding of underspecified instructions into safety-critical, dynamically feasible motion plans subject to spatiotemporal constraints. To address this challenge, we propose a unified framework that translates NL instructions into Signal Temporal Logic (STL) specifications and subsequently synthesizes trajectories via mixed-integer linear programming (MILP). Specifically, to generate executable STL formulas from free-form NL, we develop a reasoning-enhanced large language model (LLM) leveraging chain-of-thought (CoT) supervision and group-relative policy optimization (GRPO), which ensures high syntactic validity and semantic consistency. Furthermore, to resolve infeasibilities induced by stringent logical or spatial requirements, we introduce a specification repair mechanism. This module combines MILP-based diagnosis with LLM-guided semantic reasoning to selectively relax task constraints while strictly enforcing safety guarantees. Extensive simulations and real-world flight experiments demonstrate that the proposed closed-loop framework significantly improves NL-to-STL translation robustness, enabling safe, interpretable, and adaptable UAV navigation in complex scenarios.
	
\end{abstract}


\section{Introduction}
\subsection{Background and Motivation}

Low-altitude unmanned aerial vehicles (UAVs) have been increasingly deployed in mission-critical scenarios such as safety monitoring~\cite{hari2020optimal}, forest firefighting~\cite{pyq2025mllmuav}, logistics~\cite{11313542}, emergency communications~\cite{liang2025satellite}, and low-altitude networking~\cite{zhang2025toward}. Compared with high-altitude operations, low-altitude flight requires UAVs to operate in close proximity to complex urban structures, dynamic obstacles, and human activities, which imposes stringent safety and regulatory constraints. At the same time, low-altitude missions often involve high-level objectives with complex temporal, spatial, and logical requirements that must be specified clearly and executed reliably \cite{liu2023tangent}. Natural-language (NL) instructions provide an intuitive interface for expressing such high-level task intent, especially for non-expert operators. However, the inherent ambiguity and underspecification of NL stand in fundamental tension with the strict safety, temporal, and spatial requirements that low-altitude UAV navigation must satisfy.

Recent advances in large language models (LLMs) have demonstrated strong capabilities in NL understanding and high-level reasoning, which has led to growing interest in language-driven robotic autonomy~\cite{shao2025large}. Despite this progress, directly mapping NL instructions to low-level control commands remains unsuitable for safety-critical UAV operation, as such mappings lack formal guarantees, interpretability, and verifiability~\cite{tellex2020robots}. Conversely, classical model-based planning and control frameworks are capable of generating dynamically feasible trajectories under complex constraints, but they typically do not provide a systematic mechanism to interpret and enforce high-level task semantics expressed in NL~\cite{quartey2025verifiably}. This disconnect highlights the need for an intermediate representation that can faithfully capture language-level intent while remaining amenable to formal analysis and execution.

Formal methods offer a principled bridge between high-level semantic intent and low-level control. In particular, Signal Temporal Logic (STL) provides precise semantics for specifying complex temporal and spatial behaviors~\cite{belta2019formal}, enabling safety constraints and mission objectives to be expressed in a form suitable for verification and optimization-based planning. However, enabling NL UAV navigation through STL introduces two coupled challenges. First, NL instructions must be translated into STL specifications that accurately preserve the intended semantics. Second, even semantically correct STL specifications may render the underlying planning problem infeasible when temporal or spatial requirements are overly restrictive in low-altitude environments. These challenges motivate a unified framework that jointly addresses specification translation and feasibility repair, thereby ensuring both semantic fidelity and physical executability for safe and adaptable low-altitude UAV navigation.

\subsection{Related Works}

In recent years, NL-guided navigation for UAVs has attracted increasing attention. Compared with ground or indoor navigation, low-altitude UAV scenarios involve 3D motion, varying flight altitudes, and more complex spatial relations, which substantially increases the difficulty of both language understanding and navigation execution \cite{liu2023aerialvln}. Early studies mainly adopted hand-crafted grammatical rules or keyword-based mappings to convert NL commands into predefined flight actions or waypoint sequences \cite{chandarana2017fly,chandarana2017natural}. Although these methods offer interpretability and engineering controllability, their expressive capacity is limited, making them inadequate for open-vocabulary or compositional instructions and leading to poor generalization under environmental variations \cite{yao2025aeroverse}. Subsequently, deep learning techniques were widely introduced, leveraging imitation learning or reinforcement learning to jointly model NL, perception, and action spaces. This line of research gave rise to NL-guided navigation and visual language navigation (VLN) frameworks \cite{liu2023aerialvln,anderson2018vision,narasimhan2015language,wang2019reinforced,hong2021vln}. Nevertheless, their core paradigm remains task-specific policy learning, with limited capacity for explicit instruction decomposition and plan-level reasoning needed for open-ended, compositional commands.

More recently, LLM-enabled methods have pushed NL navigation toward stronger semantic understanding and flexible task execution. By supporting high-level reasoning, task decomposition, and open-vocabulary goal grounding, LLMs enable UAVs to follow longer, freer-form instructions and improve semantic adaptability \cite{zhang2025citynavagent,saxena2025uav,lee2024citynav,zhang2024interactive}. Nevertheless, many existing studies are evaluated in open simulation settings without explicitly modeling low-altitude airspace regulations, making it hard to formally verify whether a generated flight plan violates rule constraints in safety-critical 3D airspaces \cite{sanyal2025asma}.

To address these limitations, several recent studies have begun incorporating formal methods into NL-guided navigation frameworks \cite{10610123,xu2025nl2hltl2plan,wu2025selp}. Temporal logics such as linear temporal logic (LTL) and STL have been used to explicitly encode task objectives, safety constraints, and temporal requirements, which are subsequently enforced through planning algorithms. A key challenge in these frameworks is translating NL instructions into well-formed temporal-logic specifications. Early research often assumed structured or controlled language to simplify NL-to-TL mapping \cite{kress2007structured}, and some navigation-oriented approaches relied on manual or semi-automatic translation procedures to construct formulas specifying visitation order, obstacle avoidance behaviors, or timing constraints \cite{10610123}. Learning-based semantic parsing methods have subsequently been explored in several domains and have been shown to be effective in mapping NL to formal specifications \cite{gopalan2018sequence,wang2021learning,patel2020grounding}, but they typically require substantial annotated data and may generalize poorly to complex instructions that involve implicit reasoning. More recently, LLM-based methods have been explored to directly generate LTL or STL specifications from free-form NL \cite{xu2025nl2hltl2plan,wu2025selp,pan2023data,fuggitti2023nl2ltl,liu2023grounding}, offering a potential way to improve task generalization beyond what is achievable with traditional supervised parsers. Despite this progress, existing studies remain constrained by limited specification expressiveness, unstable language-to-logic mappings sensitive to prompts and context, and the fact that current LLM-based generation can also generalize poorly on complex instructions with implicit reasoning \cite{11303308,zhang2024generative}.

Moreover, the aforementioned methods often presume that human-provided NL instructions are correct and can be faithfully captured by a corresponding formal specification. In practice, the stated intent may conflict with UAV dynamical feasibility limits or low-altitude airspace constraints, rendering the synthesized plan infeasible. Once an STL specification is obtained, feasibility restoration has been extensively studied in the formal methods literature, where most approaches keep the STL structure fixed and focus on minimal parameter-level repairs. These methods typically analyze mixed-integer linear programming (MILP) encodings to identify irreducibly infeasible subsystems (IIS) or unsatisfiable cores, and restore feasibility by relaxing temporal bounds or predicate thresholds through slack variables, weighted objectives, or least-violating formulations under restricted fragments \cite{ghosh2016diagnosis,buyukkocak2022temporal,buyukkocak2025resilient,chen2023nl2tl,mao2024nl2stl}. While these optimization-driven techniques provide clear objectives and formal guarantees, they are largely language-agnostic, the decision of what to relax is typically determined by predefined costs or priorities, implicitly assuming that the original specification precisely reflects user intent. In language-grounded settings, infeasibility may instead stem from semantic ambiguity, underspecification, or NL-to-STL misinterpretation, in which case formally minimal relaxations can be semantically misaligned, such as weakening task-critical predicates when timing constraints are actually negotiable. This motivates incorporating language-level reasoning into the repair loop to guide the selection among alternative repair directions, such as relaxing temporal constraints versus predicate conditions, while still rigorously enforcing non-negotiable safety constraints through formal planning and optimization.

\subsection{Main Contributions}

This paper aims to enable safe, reliable, and interpretable low-altitude UAV navigation from natural-language (NL) instructions by jointly addressing semantic grounding, formal specification, and motion-planning feasibility. The main contributions are summarized as follows.
\begin{itemize}
	\item We develop an integrated navigation framework that converts NL instructions into STL specifications, detects planning infeasibility, and repairs the specifications based on solver feedback to restore feasibility. The framework tightly couples LLM-based semantic reasoning, STL specification translation and repair, and MILP-based motion planning within a closed-loop architecture, enabling safe and executable low-altitude UAV navigation.
	
	\item We propose an LLM-based translation method that maps NL instructions into STL specifications. The method integrates supervised chain-of-thought (CoT) alignment with group-relative policy optimization (GRPO) reinforcement learning. This pipeline improves the generation of syntactically well-formed STL specifications and increases exact-match NL-to-STL translation accuracy under strict canonical normalization.
	
	\item We introduce an LLM-assisted, systematic STL repair mechanism to handle infeasible planning instances. Diagnostic information from the MILP solver is mapped back to specific STL subformulas, and the LLM is leveraged to reason about semantic intent and prioritize repair directions, enabling selective relaxation of non-safety-critical requirements while rigorously preserving hard safety constraints.
	
	\item We validate the proposed approach through extensive simulations and real-world UAV experiments. Experimental results show that, compared with traditional NL-to-STL translation model, the proposed approach achieves higher translation accuracy while using a smaller model. In addition, the results demonstrate that the proposed closed-loop framework can generate dynamically feasible trajectories and recover from infeasible specifications.
\end{itemize}

The remainder of this paper is organized as follows. Section~II formulates the problem and reviews STL preliminaries. Section~III presents the overall framework. Sections~IV--V describe the NL-to-STL translation and the MILP-based planning and repair modules. Section~VI reports experimental results, and Section~VII concludes.

\section{Preliminaries and Problem Formulation}

In this section, we first describe the UAV dynamics and the environment representation. We then recall the STL, which serves as a formal specification layer bridging NL instructions and trajectory planning. Finally, we state the UAV NL navigation problem considered.

\subsection{UAV Dynamics and Environment Model}

We consider a UAV operating in a bounded workspace $\mathcal{W} \subset \mathbb{R}^3$ with obstacles and no-fly zones. 
The UAV state at discrete time step $k$ is defined as $\mathbf{x}_k = \begin{bmatrix} \mathbf{p}_k^\top & \mathbf{v}_k^\top \end{bmatrix}^\top$, where $\mathbf{p}_k \in \mathbb{R}^3$ and $\mathbf{v}_k \in \mathbb{R}^3$ denote the position and velocity of the UAV, respectively. The control input $\mathbf{u}_k \in \mathbb{R}^3$ corresponds to an acceleration command.

We adopt a discrete-time linear dynamics model with sampling time $\Delta t$:
\begin{equation}
	\mathbf{x}_{k+1} = \mathbf{A}\mathbf{x}_k + \mathbf{B}\mathbf{u}_k,
\end{equation}
where
\begin{equation}
	\mathbf{A} =
	\begin{bmatrix}
		\mathbf{I} & \Delta t\,\mathbf{I} \\
		\mathbf{0} & \mathbf{I}
	\end{bmatrix},
	\qquad
	\mathbf{B} =
	\begin{bmatrix}
		\frac{1}{2}\Delta t^2\,\mathbf{I} \\
		\Delta t\,\mathbf{I}
	\end{bmatrix}.
\end{equation}
Here, $\mathbf{I}$ denotes the identity matrix of appropriate dimension and $\mathbf{0}$ denotes a zero matrix. State and control constraints are imposed as:
\begin{equation}
	\mathbf{x}_k \in \mathcal{X}, \qquad \mathbf{u}_k \in \mathcal{U},
\end{equation}
where $\mathcal{X}$ and $\mathcal{U}$ are convex polyhedral sets encoding workspace bounds, velocity limits, and actuation constraints.

The environment contains a set of labeled regions $\{\mathcal{R}_i\}_{i=1}^M$ that represent task-relevant areas in the workspace. Obstacles and no-fly zones are modeled as forbidden regions that must be avoided by the UAV for all the $k$ time steps.

\begin{figure*}[t]
	\centering
	\includegraphics[width=1.0\linewidth]{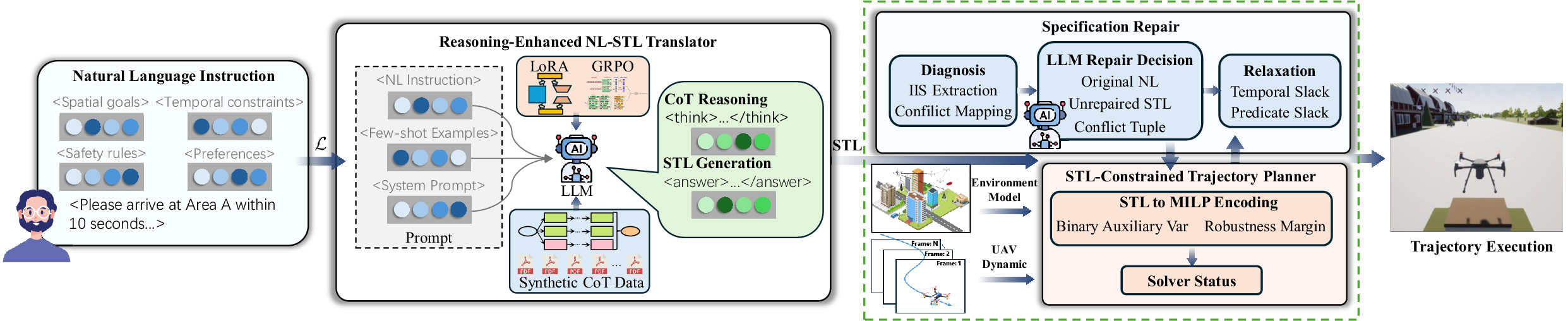}
	\caption{Overview of the proposed language-guided planning framework.
		A natural-language instruction is translated into an STL specification, which is enforced by a constrained planner.
		When infeasibility is detected, solver feedback is used to guide specification refinement and re-planning in a closed loop.}
	\label{fig:system_overview}
\end{figure*}
\subsection{Signal Temporal Logic Specifications}
\label{subsec:stl}

STL has been extensively used to specify and verify temporal properties of dynamical systems. In this work, STL formulas are interpreted over discrete-time UAV state trajectories $\mathbf{x}_{0:H} = \{\mathbf{x}_0, \mathbf{x}_1, \dots, \mathbf{x}_H\}$, where $H \in \mathbb{N}$ denotes the planning horizon.

An STL formula $\varphi$ is defined recursively as:
\begin{equation}
	\begin{split}
		\varphi ::=\;& \top \mid \mu \mid \neg \varphi \mid \varphi_1 \wedge \varphi_2 \mid \varphi_1 \vee \varphi_2 \\
		&\mid \mathcal{G}_{[a,b]} \varphi \mid \mathcal{F}_{[a,b]} \varphi \mid \varphi_1 \mathcal{U}_{[a,b]} \varphi_2,
	\end{split}
\end{equation}
where $a,b \in \mathbb{N}$ with $a \le b$ denote discrete-time bounds.Here, the temporal operators $\mathcal{F}$, $\mathcal{G}$, and $\mathcal{U}$ correspond to the \textit{eventually}, \textit{always}, and \textit{until} operators, respectively. And $\varphi_1$ and $\varphi_2$ denote arbitrary STL subformulas. The atomic predicate $\mu$ is defined as an inequality over the system state:
\begin{equation}
	\mu(\mathbf{x}_k) = g(\mathbf{x}_k) \ge 0,
\end{equation}
where $g(\cdot)$ is an affine function of the system state. Such predicates can encode region membership, obstacle clearance, no-fly zones, kinematic limits, and other safety envelopes.

The STL semantics is recursively defined as follows \cite{belta2019formal}:
\begin{align}
	% (7) 简短，保持原样
	(\mathbf{x}_{0:H}, k) \models \mu
	&\iff g(\mathbf{x}_k) \ge 0, \\
	% (8) 简短，保持原样
	(\mathbf{x}_{0:H}, k) \models \neg \varphi
	&\iff (\mathbf{x}_{0:H}, k) \not\models \varphi, \\
	% (9) 在 \wedge 处换行
	(\mathbf{x}_{0:H}, k) \models \varphi_1 \wedge \varphi_2
	&\iff (\mathbf{x}_{0:H}, k) \models \varphi_1 \notag \\
	&\quad\; \wedge (\mathbf{x}_{0:H}, k) \models \varphi_2, \\
	% (10) 在 \vee 处换行
	(\mathbf{x}_{0:H}, k) \models \varphi_1 \vee \varphi_2
	&\iff (\mathbf{x}_{0:H}, k) \models \varphi_1 \notag \\
	&\quad\; \vee (\mathbf{x}_{0:H}, k) \models \varphi_2, \\
	% (11) 在量词后换行
	(\mathbf{x}_{0:H}, k) \models \mathcal{F}_{[a,b]} \varphi
	&\iff \exists k' \in [k+a, k+b] : \notag \\
	&\quad\; (\mathbf{x}_{0:H}, k') \models \varphi, \\
	% (12) 在量词后换行
	(\mathbf{x}_{0:H}, k) \models \mathcal{G}_{[a,b]} \varphi
	&\iff \forall k' \in [k+a, k+b] : \notag \\
	&\quad\; (\mathbf{x}_{0:H}, k') \models \varphi, \\
	% (13-14) Already split, refined alignment
	(\mathbf{x}_{0:H}, k) \models \varphi_1 \mathcal{U}_{[a,b]} \varphi_2
	&\iff \exists k' \in [k+a, k+b] \text{ s.t. } \notag \\
	&\quad\; (\mathbf{x}_{0:H}, k') \models \varphi_2 \notag \\
	&\quad\; \wedge \forall k'' \in [k, k'] : (\mathbf{x}_{0:H}, k'') \models \varphi_1.
\end{align}

STL also defines a robust semantics by associating each formula $\varphi$ with a real-valued function $\rho_\varphi(\mathbf{x}_{0:H}, k)$ such that
$(\mathbf{x}_{0:H}, k) \models \varphi$ if and only if $\rho_\varphi(\mathbf{x}_{0:H}, k) \ge 0$.
The magnitude of $\rho_\varphi$ can be interpreted as the margin by which $\varphi$ is satisfied or violated.
Following robust STL semantics in \cite{donze2010robust}, the quantitative semantics is defined recursively as:
\begin{align}
	\rho_\mu(\mathbf{x}_{0:H}, k) &= g(\mathbf{x}_k), \\
	\rho_{\neg \varphi}(\mathbf{x}_{0:H}, k) &= -\rho_{\varphi}(\mathbf{x}_{0:H}, k), \\
	\rho_{\varphi_1 \wedge \varphi_2}(\mathbf{x}_{0:H}, k) &= \min\Bigl(\rho_{\varphi_1}(\mathbf{x}_{0:H}, k),\, \rho_{\varphi_2}(\mathbf{x}_{0:H}, k)\Bigr), \\
	\rho_{\varphi_1 \vee \varphi_2}(\mathbf{x}_{0:H}, k) &= \max\Bigl(\rho_{\varphi_1}(\mathbf{x}_{0:H}, k),\, \rho_{\varphi_2}(\mathbf{x}_{0:H}, k)\Bigr), \\
	\rho_{\mathcal{G}_{[a,b]}\varphi}(\mathbf{x}_{0:H}, k) &= \min_{k' \in [k+a,\, k+b]} \rho_{\varphi}(\mathbf{x}_{0:H}, k'), \\
	\rho_{\mathcal{F}_{[a,b]}\varphi}(\mathbf{x}_{0:H}, k) &= \max_{k' \in [k+a,\, k+b]} \rho_{\varphi}(\mathbf{x}_{0:H}, k'), \\
	\rho_{\varphi_1 \mathcal{U}_{[a,b]} \varphi_2}(\mathbf{x}_{0:H}, k) &= \max_{k' \in [k+a,\, k+b]} \min\Bigl( \rho_{\varphi_2}(\mathbf{x}_{0:H}, k'), \notag \\
	&\qquad\qquad \min_{k'' \in [k,\, k']} \rho_{\varphi_1}(\mathbf{x}_{0:H}, k'') \Bigr).
\end{align}

In this paper, STL provides a verifiable intermediate specification between NL instructions and optimization-based trajectory planning. Safety-critical requirements are enforced as hard STL constraints. Task-related objectives are expressed via temporal operators and may be selectively relaxed when the resulting STL-constrained planning problem is infeasible.

\subsection{Problem Definition}

Consider a UAV operating in a known environment with discrete-time dynamics and admissible state and control sets. Let $\mathcal{L}$ denote a NL instruction describing a navigation task. The instruction is interpreted as a STL specification:
\begin{equation}
	\varphi = T(\mathcal{L}),
\end{equation}
where $T(\cdot)$ denotes a translation model.

Given the induced STL specification $\varphi$ and a finite planning horizon $H$, the objective is to compute a dynamically feasible state trajectory $\mathbf{x}_{0:H}$ and control sequence $\mathbf{u}_{0:H-1}$ such that:
\begin{equation}
	\mathbf{x}_{0:H} \models \varphi .
\end{equation}
The resulting trajectory must satisfy the system dynamics, state and control constraints, and all safety-critical requirements encoded in $\varphi$.

The overall NL navigation problem addressed in this paper can be summarized by the following relation:
\begin{equation}
	(\mathcal{L}, H) \ \mapsto\  (\varphi, \mathbf{x}_{0:H}, \mathbf{u}_{0:H-1}).
\end{equation}

Due to ambiguity in natural language or conflicts among temporal and spatial task requirements, the STL-constrained planning problem induced by $\varphi$ may be infeasible.
In such cases, the goal is to restore feasibility by minimally modifying task-related components of the specification while strictly preserving all safety-critical constraints.

\section{System Overview and Framework}
\label{sec:system_overview}

Fig.~\ref{fig:system_overview} shows the proposed framework for low-altitude UAV navigation from a NL instruction $\mathcal{L}$. The system translates $\mathcal{L}$ into a verifiable STL specification, plans a dynamically feasible trajectory under STL and physical constraints, and triggers specification repair when the planning problem is infeasible.

\emph{Reasoning-Enhanced NL-STL Translator.}
Given $\mathcal{L}$ , the translator uses a structured prompt and a reasoning-enhanced LLM to generate an STL specification $\varphi$. The model is trained on a synthetic NL-STL dataset augmented with CoT reasoning traces, and is further fine-tuned using parameter-efficient LoRA and GRPO to improve STL syntactic validity and semantic consistency.

\emph{STL-Constrained Trajectory Planner.}
The planner converts $\varphi$ into a MILP by introducing binary satisfaction variables for STL subformulas and a robustness margin variable. The MILP also incorporates the environment model and UAV dynamics with state/control limits. If the solver is feasible, it returns a trajectory for execution, otherwise, it outputs an infeasibility status that activates repair.

\emph{Specification Repair.}
Upon infeasibility, the system extracts an IIS and maps conflicting MILP constraints back to STL subformulas and time indices to form a conflict tuple. Using the original NL instruction, unrepaired STL, and conflict tuple, an LLM selects a repair mode for each conflict: \emph{temporal relaxation} or \emph{predicate relaxation}, while safety constraints are never relaxed. The selected mode is implemented via temporal or predicate slack variables with penalties, and the repaired STL is reconstructed and sent back to the planner for re-solving.

\begin{figure*}[t]
	\centering
	\includegraphics[width=0.9\textwidth]{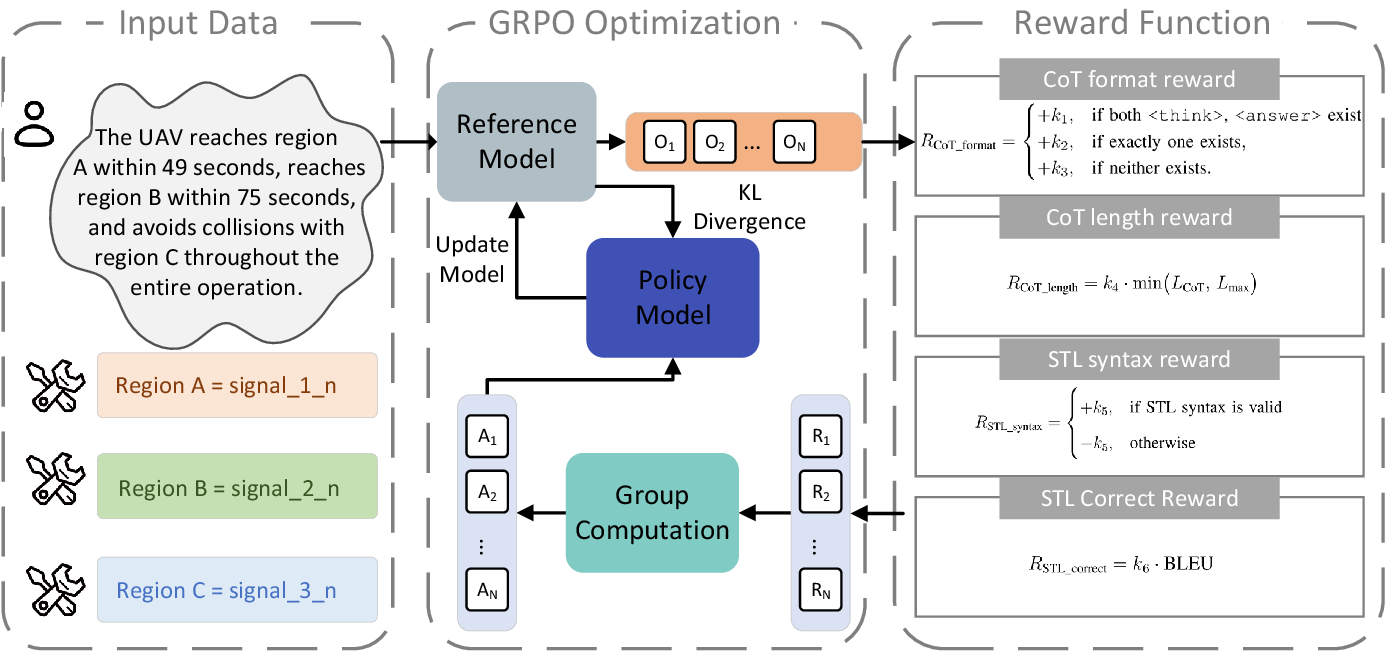}
	\caption{GRPO-based RL framework for reasoning-enhanced NL-to-STL generation.}
	\label{fig:grpo}
\end{figure*}

Overall, the framework realizes $\mathcal{L}\!\rightarrow\!\varphi\!\rightarrow\!(\mathbf{x}_{0:H},\mathbf{u}_{0:H-1})$ with automatic diagnosis-and-repair in the loop, enabling safe and executable UAV navigation while keeping repairs interpretable and minimally task-degrading.

\section{Reasoning-Enhanced NL to STL Translation}

This section presents a reasoning-enhanced pipeline for translating NL instructions into STL specifications. We construct training data by augmenting an existing NL-to-STL dataset with explicit reasoning traces. A CoT data generation pipeline generates intermediate structured reasoning to connect each instruction to its target STL formula, forming an NL-CoT-STL corpus. We then use this corpus for cold start supervised fine-tuning and apply GRPO reinforcement learning to further improve STL syntactic correctness and semantic consistency. The overall training pipeline, including cold-start SFT and GRPO-based reinforcement learning, is illustrated in Fig.~\ref{fig:grpo}.

\subsection{Synthetic Dataset and CoT Data Generation}

\begin{figure}[h]
	\centering
	\includegraphics[width=0.40\textwidth]{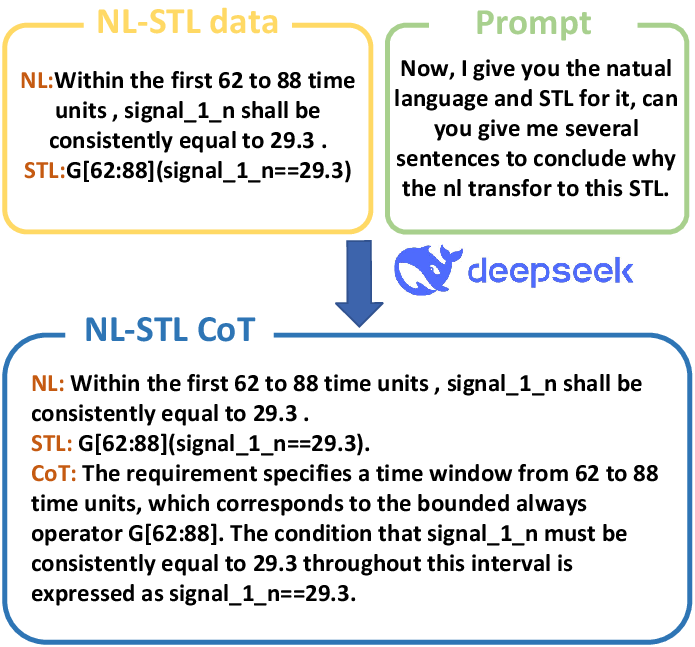}
	\caption{CoT Data Engine for augmenting NL2TL with intermediate reasoning traces.}
	\label{fig:cot}
\end{figure}

To improve NL-to-STL translation, we build a synthetic training corpus based on NL2TL \cite{chen2023nl2tl}. We adopt NL2TL as the base dataset and collect 20K NL-STL pairs from it. While NL2TL provides aligned NL instructions and STL formulas, the intermediate semantic decomposition from NL to STL is missing, making it difficult for the model to learn how linguistic cues are mapped to STL operators and temporal constraints. To address this issue, each NL-STL pair is augmented with an intermediate CoT, which serves as a structured bridge between the NL instruction and the corresponding STL specification. The CoT captures the essential semantic decomposition steps required for temporal logic construction, transforming each original NL-STL pair into a NL-CoT-STL triplet. The overall data generation pipeline is illustrated in Fig.~\ref{fig:cot}.

The CoT annotations are generated using DeepSeek-V3.1 \cite{deepseekmath}. Given a NL instruction and its ground-truth STL formula, the model reconstructs the key reasoning steps required for temporal logic construction, including predicate identification, temporal bound extraction, operator selection, and formula composition. These CoT traces serve as an explicit intermediate representation that exposes the semantic structure underlying the NL-to-STL mapping.

Applying this pipeline to the selected NL2TL samples yields a synthetic dataset with explicit supervision over both reasoning and final STL outputs. This dataset forms the basis for the subsequent cold-start supervised fine-tuning stage and improves the structural robustness and semantic consistency of downstream STL generation.

\subsection{Cold Start Stage}
﻿
Recent advances in reinforcement learning for reasoning, such as DeepSeek-R1 \cite{deepseekr1}, indicate that policy optimization can improve long-horizon reasoning. However, NL-to-STL translation requires both explicit CoT reasoning and strict adherence to STL syntax. Acquiring these capabilities purely through end-to-end reinforcement learning from scratch is often unstable, due to rigid syntactic constraints and sparse task-level rewards. Therefore, we introduce a cold-start supervised fine-tuning (SFT) stage to initialize the model before GRPO training.
﻿

The cold-start stage targets format alignment and reasoning induction. Given the rigidity of STL syntax and the scarcity of temporal-logic structures in general pre-training corpora, prompt-based approaches alone are insufficient to guarantee consistently well-formed outputs. We fine-tune the model on the synthetic NL-CoT-STL dataset constructed in the previous section, where each sample follows a structured output pattern: the intermediate reasoning is enclosed within \texttt{<think>} tags, and the final STL specification is enclosed within \texttt{<answer>} tags. We optimize the model using a standard maximum-likelihood objective, and implement SFT via parameter-efficient fine-tuning with Low-Rank Adaptation (LoRA) \cite{hu2022lora} to reduce memory and computational overhead while preserving the base model's general language capability.
﻿

This cold start initialization encourages the model to produce coherent reasoning traces and syntactically valid STL formulas in a stable and controllable manner. By reducing the search space and variance encountered in subsequent reinforcement learning, it enables the GRPO optimization stage to focus on improving global structural validity and semantic fidelity, rather than correcting low-level formatting errors.
﻿

\subsection{GRPO Reinforcement Learning}

Although SFT enables the model to imitate correct outputs, it optimizes token-level likelihood rather than global structural validity. For structured outputs such as STL formulas, minor token errors can invalidate the entire formulas.

Reinforcement learning addresses this issue by directly optimizing task-level rewards. While Proximal Policy Optimization (PPO) \cite{PPO} or Group-based Proximal Policy Optimization (GPPO) \cite{zhang2025covert} are a widely adopted RL method for LLMs, it requires an additional value network and often suffers from high computational cost and training instability. To overcome these limitations, we adopt GRPO \cite{deepseekmath}. GRPO eliminates the need for a value network by estimating the baseline using group-level reward statistics.

We explicitly structure the model output into a reasoning process enclosed by \texttt{<think>} tags and a final STL formula enclosed by \texttt{<answer>} tags. The overall RL pipeline follows three stages including policy sampling, reward computation, and policy update.

Given an input instruction, we sample a group of $G$ candidate outputs $\{O_i\}_{i=1}^G$ from the current policy $\pi_{\theta_{\text{old}}}$ with a relatively high temperature coefficient $\tau$ to encourage exploration and output diversity. For each sampled output $O_i$, we compute a composite reward by summing four components:
\begin{equation}
	R_i =
	R_{\text{CoT\_format}}
	+ R_{\text{CoT\_length}}
	+ R_{\text{STL\_syntax}}
	+ R_{\text{STL\_correct}}.
\end{equation}
Each reward term is computed from the sampled output $O_i$.

To encourage the structural integrity of the CoT format, we define the \textbf{CoT format reward} with three cases:
\begin{equation}
	R_{\text{CoT\_format}} =
	\begin{cases}
		+k_1, & \text{if both \texttt{<think>}, \texttt{<answer>}} \text{ exist}\\
		+k_2, & \text{if exactly one exists},\\
		+k_3, & \text{if neither exists}.
	\end{cases}
\end{equation}

To encourage sufficient reasoning while penalizing verbosity, we define the \textbf{CoT length reward}:
\begin{equation}
	R_{\text{CoT\_length}} = k_4 \cdot \min\!\big(L_{\text{CoT}},\, L_{\text{max}}\big),
\end{equation}
where $L_{\text{CoT}}$ is the number of tokens inside the \texttt{<think>} span and $L_{\text{max}}$ is the maximum allowed CoT length.

To enforce syntactic validity of the generated STL, we define the \textbf{STL syntax reward}:
\begin{equation}
	R_{\text{STL\_syntax}} =
	\begin{cases}
		+k_5, & \text{if STL syntax is valid}\\
		-k_5, & \text{otherwise}
	\end{cases}
\end{equation}
where validity requires that all variables belong to the predefined set and that all temporal operators are structurally complete.

To provide a dense similarity-based signal, we define a \textbf{BLEU-based STL correctness reward}:
\begin{equation}
	R_{\text{STL\_correct}} = k_6 \cdot \text{BLEU},
\end{equation}
where BLEU \cite{papineni2002bleu} is computed by combining $n$-gram precision and a brevity penalty:
\begin{equation}
	\text{BLEU} = \text{BP} \cdot \exp\left(\sum_{n=1}^{N} w_n \log p_n\right).
\end{equation}
Here, $p_n$ is the modified $n$-gram precision of order $n$ between the generated and reference STL sequences, $w_n$ is the corresponding weight with $\sum_{n=1}^{N} w_n = 1$, and $\text{BP}$ is the brevity penalty determined by the lengths of the generated and reference sequences. We use $\{k_j\}_{j=1}^{6}$ as scalar reward coefficients to balance the contributions of the reward terms.

After reward computation, we perform policy update using GRPO. For each input, the rewards $\mathbf{R}=\{R_i\}_{i=1}^G$ of the sampled group are normalized relative to the group mean and variance, yielding the group-relative advantage:
\begin{equation}
	\hat{A}_{i,t} = \frac{R_i - \text{mean}(\mathbf{R})}{\text{std}(\mathbf{R})},
\end{equation}
where $t\in\{1,\dots,|O_i|\}$ indexes tokens in the generated trajectory $O_i$. The same group-relative advantage $\hat A_{i,t}$ is assigned to all tokens in the trajectory $O_i$.

The GRPO objective combines a clipped policy update with KL regularization against a reference policy:
\begin{equation}
	\begin{aligned}
		L_{\text{GRPO}}(\theta)
		&= \mathbb{E}\Bigg[
		\frac{1}{G} \sum_{i=1}^G \frac{1}{|O_i|}
		\sum_{t=1}^{|O_i|}
		\Big(
		L_{\text{CLIP}}(\theta, \hat{A}_{i,t}) \\
		&\quad - \beta\, \mathbb{D}_{\text{KL}}
		\big[\pi_{\theta} \,\|\, \pi_{\text{ref}}\big]
		\Big)
		\Bigg].
	\end{aligned}
\end{equation}

For each token position $t$ in the trajectory $O_i$, we compute the importance ratio:
\begin{equation}
	r_{i,t}(\theta)=\dfrac{\pi_\theta(a_{i,t}\mid s_{i,t})}{\pi_{\theta_{\text{old}}}(a_{i,t}\mid s_{i,t})},
\end{equation}
where $a_{i,t}$ denotes the $t$-th generated token in trajectory $O_i$, and $s_{i,t}$ denotes the corresponding generation context, consisting of the input and the previously generated tokens.

We then define the clipped surrogate objective:
\begin{equation}
	L_{\mathrm{CLIP}}^{(i,t)}=\min\!\Big(r_{i,t}(\theta)\hat A_{i,t},~\text{clip}(r_{i,t}(\theta),1-\epsilon,1+\epsilon)\hat A_{i,t}\Big).
\end{equation}
The policy parameters are then updated by gradient ascent on $L_\text{GRPO}(\theta)$.

The complete GRPO training procedure, including cold-start initialization and main reinforcement learning phases, is summarized in Algorithm~\ref{al1}.

\begin{algorithm}[H]
	\caption{GRPO Training for NL-to-STL Translator with Chain-of-Thought}
	\label{al1}
	\begin{algorithmic}[1]
		
		\STATE \textbf{Phase 0: Cold Start}
		\STATE Build $\mathcal{D}$ of NL-CoT-STL triplets with \texttt{<think>} and \texttt{<answer>} tags; SFT to obtain $\pi_\theta^{(0)}$;
		\STATE Initialize $\pi_\theta \leftarrow \pi_\theta^{(0)}$ and reference policy $\pi_{\text{ref}} \leftarrow \pi_\theta^{(0)}$;
		
		\STATE \textbf{Phase 1: GRPO Main Training}
		\FOR{$k=1$ \textbf{to} $K$}
		\STATE Set $\pi_{\theta_{\text{old}}}\leftarrow \pi_\theta$;
		\STATE Sample a group $\{O_i\}_{i=1}^G$ from $\pi_{\theta_{\text{old}}}(\cdot\mid x)$ with temperature $\tau$;
		\STATE Compute rewards $\{R_i\}_{i=1}^G$ where $R_i=R_{\text{CoT\_format}}+R_{\text{CoT\_length}}+R_{\text{STL\_syntax}}+R_{\text{STL\_correct}}$;
		\STATE Compute $\mu=\mathrm{mean}(\{R_i\})$, $\sigma=\mathrm{std}(\{R_i\})$, and set $\hat A_{i,t}=(R_i-\mu)/(\sigma+\varepsilon)$ for all $t=1,\dots,|O_i|$;
		
		\STATE Compute $L_{\text{GRPO}}(\theta)=\frac{1}{G}\sum_{i=1}^G \frac{1}{|O_i|}\sum_{t=1}^{|O_i|}\Big(L_{\mathrm{CLIP}}^{(i,t)}-\beta\,\mathbb{D}_{\mathrm{KL}}[\pi_\theta\|\pi_{\text{ref}}]\Big)$, where
		\STATE \hspace{1.6em}$r_{i,t}(\theta)=\pi_\theta(a_{i,t}\!\mid\! s_{i,t})/\pi_{\theta_{\text{old}}}(a_{i,t}\!\mid\! s_{i,t})$ and
		\STATE \hspace{1.6em}$L_{\mathrm{CLIP}}^{(i,t)}=\min\!\Big(r_{i,t}(\theta)\hat A_{i,t},~\text{clip}(r_{i,t}(\theta),1-\epsilon,1+\epsilon)\hat A_{i,t}\Big)$;
		
		\STATE Update by gradient ascent: $\theta \leftarrow \theta + \eta \nabla_\theta L_{\text{GRPO}}(\theta)$;
		\ENDFOR
		
	\end{algorithmic}
\end{algorithm}

\section{STL-Constrained Trajectory Planning and Specification Repair}
\label{sec:stl_planning_and_repair}

Given the discrete-time UAV dynamics and the STL specification generated from NL instructions, we formulate the trajectory generation problem as a constrained planning problem over a finite horizon.
This section presents a unified formulation that integrates STL-constrained trajectory planning with feasibility diagnosis and specification repair.

\subsection{MILP Encoding of STL Satisfaction}
\label{subsec:milp_encoding}

Given the discrete-time UAV dynamics and the STL specification $\varphi$ translated from the NL instruction by the LLM, we encode satisfaction of $\varphi$ over a finite planning horizon $H$ using a MILP formulation.

For each STL subformula $\psi$ and discrete time step $k$, we introduce a binary auxiliary variable $z_{\psi,k}\in\{0,1\}$ indicating whether $\psi$ is
satisfied at time $k$. The atomic predicate $\mu$ is defined as an inequality over the system state $\mathbf{x}_k$ of the form $\mu(\mathbf{x}_k) \ge 0$.
The implication between the binary variable and predicate satisfaction is encoded using Big-$M$ constraints as
\begin{equation}
	\begin{aligned}
		\mu(\mathbf{x}_k) + (1 - z_{\mu,k}) M &\ge \gamma, \quad\mu(\mathbf{x}_k) - z_{\mu,k} M &\le \gamma,
	\end{aligned}
	\label{eq:bigm_predicate}
\end{equation}
where $M>0$ is a sufficiently large constant and $\gamma \ge 0$ is a global robustness margin variable.
Unlike the quantitative STL robustness $\rho_\varphi(\mathbf{x}_{0:H},k)$ defined in Section~II-B, which evaluates the satisfaction margin of a formula on a given trajectory, $\gamma$ is an optimization variable that enforces a uniform lower bound on predicate satisfaction across the entire trajectory.

Boolean compositions of STL formulas are encoded recursively.
For a conjunction $\psi=\bigwedge_i \psi_i$, we impose:
\begin{equation}
	z_{\psi,k} \le z_{\psi_i,k+\Delta_i}, \quad \forall i,
	\label{eq:and_encoding}
\end{equation}
and for a disjunction $\psi=\bigvee_i \psi_i$:
\begin{equation}
	z_{\psi,k} \le \sum_i z_{\psi_i,k+\Delta_i},
	\label{eq:or_encoding}
\end{equation}
where $\Delta_i$ denotes the relative time offset induced by the syntax tree of $\psi$.

Temporal operators are unfolded over their discrete-time intervals according to the STL semantics defined in Section~II-B.
Specifically, $\mathcal{G}_{[a,b]}$ operators are expanded as conjunctions over $\{k+a,\dots,k+b\}$, while $\mathcal{F}_{[a,b]}$ operators are expanded as disjunctions over the same interval. The until operator $\mathcal{U}_{[a,b]}$ is handled analogously using its standard Boolean expansion. All resulting Boolean constraints are encoded using Equations \eqref{eq:and_encoding}--\eqref{eq:or_encoding}.

Satisfaction of the overall STL specification is enforced by introducing a root
variable $z_{\varphi,0}$ and requiring:
\begin{equation}
	z_{\varphi,0} = 1, \qquad \gamma \ge 0.
	\label{eq:root_satisfaction}
\end{equation}
Any feasible solution to the resulting MILP therefore corresponds to a
trajectory $\mathbf{x}_{0:H}$ that satisfies $\varphi$.

\subsection{STL-Constrained Optimization Formulation}
\label{subsec:stl_optimization}

Combining the system dynamics, state and control constraints, and the MILP encoding of STL satisfaction, we obtain the following mixed-integer optimization problem over the horizon $H$:
\begin{subequations}\label{prob:P1}
	\begin{align}
		\tag{P1}\label{prob:P1_tag}
		\operatorname*{min.}_{\mathbf{x}_{0:H},\,\mathbf{u}_{0:H-1},\,\mathbf{z},\,\gamma}\quad
		& -\gamma + \sum_{k=0}^{H-1}\left(\mathbf{x}_k^\top \mathbf{Q}\mathbf{x}_k + \mathbf{u}_k^\top \mathbf{R}\mathbf{u}_k\right) \\
		\text{s.t.}\quad
		& \mathbf{x}_0 = \mathbf{x}_{\text{fixed}},\\
		& \mathbf{x}_{k+1} = \mathbf{A}\mathbf{x}_k + \mathbf{B}\mathbf{u}_k, \forall k=0,\dots,H-1,\\
		& \mathbf{x}_k \in \mathcal{X}, \mathbf{u}_k \in \mathcal{U},\quad \forall k=0,\dots,H-1,\\
		& \eqref{eq:bigm_predicate}-\eqref{eq:root_satisfaction}.\nonumber
	\end{align}
\end{subequations}

The objective in Problem~(P1) seeks a trajectory that maximizes the global robustness margin $\gamma$ while penalizing state deviation and control effort
through quadratic regularization. Feasibility with $\gamma \ge 0$ guarantees satisfaction of the STL specification, and larger values of $\gamma$ correspond
to increased robustness against perturbations.

\subsection{Feasibility Certification and IIS Extraction}
\label{subsec:iis}

If the solver declares  Problem~(P1) infeasible, we perform feasibility diagnosis using an IIS, which identifies a minimal set of constraints and variable bounds that cannot be satisfied simultaneously. We adopt the IIS-based diagnosis procedure in \cite{ghosh2016diagnosis} to localize infeasibility at the specification level.

To enable semantic interpretation of solver feedback, each STL-induced constraint in the MILP is associated with a traceability record that links it to the originating STL subformula and time index. Using this mapping, the constraints contained in the IIS are projected back to the STL syntax tree and summarized as a set of infeasible atomic events, each characterized by an atomic predicate and its associated temporal context. This diagnosis result localizes the source of infeasibility in a form suitable for high-level reasoning and serves as structured input to the subsequent specification repair stage.

\subsection{LLM-Guided Repair via Predicate-Temporal Choice}
\label{subsec:repair_slack}

When the STL-constrained optimization problem is infeasible, our objective is to restore feasibility by minimally modifying task-related requirements while strictly preserving all safety-critical constraints.
Unlike prior approaches that require the designer to manually specify which predicates or temporal parameters are eligible for repair, we delegate the choice of \emph{repair dimension} to a LLM, which operates at the semantic level and does not directly manipulate numerical optimization variables.

For each infeasible atomic event identified by the IIS-based diagnosis in Section~\ref{subsec:iis}, the LLM is provided with a structured input that combines symbolic, temporal, and semantic information.
Specifically, the input consists of: (i) the original NL instruction from which the task specification was generated, (ii) the unrepaired STL specification $\varphi$, and (iii) a structured description of each diagnosed atomic event represented as a tuple $(\mu,\sigma,\mathcal{O},\mathcal{R})$, where $\mu$ denotes the atomic predicate inequality, $\sigma$ is its discrete-time support interval induced by the STL semantics, $\mathcal{O}$ is the parent temporal operator in the STL syntax tree, and $\mathcal{R}$ characterizes the semantic role of the predicate.
This representation allows the LLM to reason jointly over the original task intent, the formal specification structure, and the localized infeasibility information, without exposing solver-level variables or numerical relaxation parameters.

Safety-critical predicates, such as obstacle avoidance and no-fly constraints, are explicitly excluded from this interface and are never considered for relaxation.

Given this input, the LLM outputs a binary decision selecting exactly one admissible repair mode for each infeasible atomic event: \emph{predicate relaxation} or \emph{temporal relaxation}.
To preserve interpretability and a clear separation between spatial and temporal modifications, mixed repairs for a single atomic event are intentionally disallowed.
The LLM does not propose relaxation magnitudes, threshold values, or modified temporal bounds; it only determines which repair dimension is permitted, while the quantitative extent of repair is determined entirely by the optimization layer.

The LLM's decisions are then realized at the numerical level through selective relaxation in the MILP.
If predicate relaxation is selected for an atomic event, a nonnegative slack variable $s_{\mu,k} \ge 0$ is introduced to relax the corresponding Big-$M$ constraint:
\begin{equation}
	\mu(\mathbf{x}_k) + (1 - z_{\mu,k})M + s_{\mu,k} \ge \gamma.
	\label{eq:predicate_slack}
\end{equation}
If temporal relaxation is selected, nonnegative temporal slack variables $\tau_{\psi,k}$ are introduced to relax the Boolean constraints arising from temporal operator expansion.
For atomic events deemed non-negotiable, no relaxation variables are introduced and the corresponding constraints remain unchanged.

After introducing the selected relaxations, we resolve the optimization problem with an augmented objective that penalizes the total relaxation magnitude:
\begin{equation}
	\operatorname{min.} \;\; -\gamma
	+ \sum_{k}\!\left(\mathbf{x}_k^\top \mathbf{Q}\mathbf{x}_k
	+ \mathbf{u}_k^\top \mathbf{R}\mathbf{u}_k\right)
	+ \lambda_p \sum s_{\mu,k}
	+ \lambda_t \sum \tau_{\psi,k},
	\label{eq:repair_objective}
\end{equation}
where $\lambda_p,\lambda_t>0$ weight the relative cost of predicate and temporal relaxations.
The optimized relaxation values provide quantitative guidance for constructing an explicit repaired STL specification $\tilde{\varphi}$: predicate relaxations are translated into adjusted numerical thresholds or geometric parameters, while temporal relaxations are mapped to modified temporal bounds. Finally, $\tilde{\varphi}$ is reconstructed without slack variables and re-encoded into a standard STL-constrained MILP; solving this problem yields a dynamically feasible trajectory that satisfies $\tilde{\varphi}$ exactly.

The introduction of the LLM does not modify the underlying IIS-based diagnosis or the optimization-based repair mechanism.
For any fixed set of LLM repair-mode selections, feasibility and quantitative minimality are enforced by the MILP formulation, while the LLM influences only the admissible repair dimension for each diagnosed atomic event.

\section{Experimental Evaluation}
\label{sec:experiments}

This section presents a comprehensive experimental evaluation of the proposed NL navigation framework. The experiments are designed to assess the effectiveness of the system at multiple levels, ranging from NL understanding and formal specification generation to closed-loop trajectory planning and execution. We first evaluate the performance of the proposed reasoning-enhanced NL-to-STL translation model in isolation, focusing on its syntactic correctness with respect to ground-truth temporal logic specifications. We then assess the full planning framework in simulation, where the translated STL specifications are integrated with the MILP-based trajectory planner. These experiments evaluate the system's ability to generate dynamically feasible trajectories, enforce safety-critical constraints, and recover from infeasible specifications through selective repair. Finally, we demonstrate the applicability of the proposed approach in real-world experiments using a quadrotor UAV platform in a representative search-and-rescue task.

\subsection{NL-to-STL Translation Experiments}
\label{subsec:nl2stl_experiments}

This subsection evaluates the proposed reasoning-enhanced NL to STL Translation model, with an emphasis on its ability to generate syntactically valid and semantically accurate STL specifications from NL instructions. The translator is built upon \textbf{Qwen2.5-0.5B-Instruct} and trained using the reasoning-enhanced pipeline described in Section~IV.

All experiments are conducted on the NL2TL dataset~\cite{chen2023nl2tl}, where 20K NL-STL pairs are used for training and an additional 4K samples are held out for testing. We compare against several representative baselines, including a Llama2-finetuned model \cite{mao2024nl2stl}, a T5-finetuned model \cite{chen2023nl2tl}, the original Qwen2.5-0.5B-Instruct model without task-specific adaptation, and a Qwen2.5-0.5B-Instruct model fine-tuned on NL-STL pairs without CoT supervision. All fine-tuned baselines are trained on the same data split and evaluated under identical decoding constraints.

Translation performance is measured using accuracy, defined as the exact-match rate between the generated STL specification and the ground-truth formula after canonical normalization of syntax and operators.This metric is intentionally strict, as any syntactic error, incorrect temporal bound, or logical mismatch renders the output incorrect, reflecting the requirement that generated specifications must be directly executable by downstream formal planners.

\begin{figure}[h]
	\centering
	\includegraphics[width=0.4\textwidth,trim=13 10 10 10,	clip]{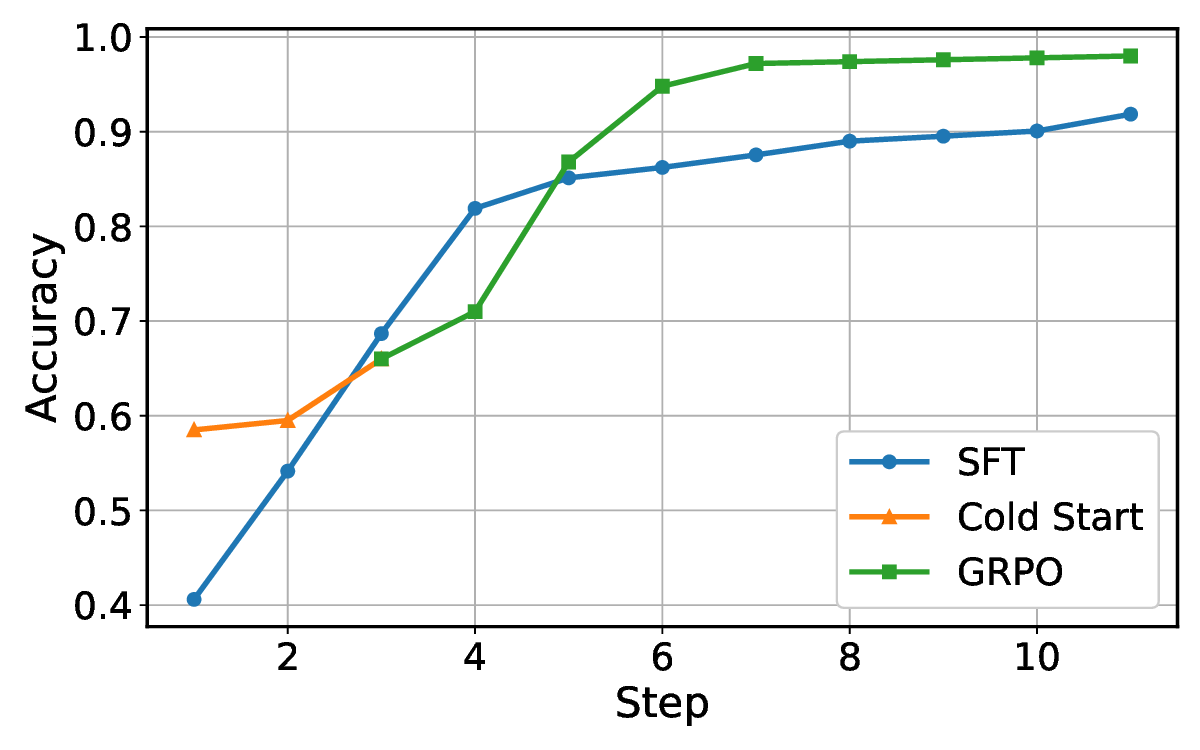}
	\caption{NL-to-STL translation accuracy during SFT and GRPO training.}
	\label{fig:nl2stl_convergence}
\end{figure}

Fig.~\ref{fig:nl2stl_convergence} shows the convergence of the proposed NL-to-STL translation model across training stages. SFT denotes supervised fine-tuning on data without CoT. The accuracy improves steadily as the model learns the rigid STL grammar and its alignment with natural-language expressions, but the gains gradually saturate. The cold-start stage corresponds to supervised fine-tuning on data with CoT, which yields a clear jump in accuracy, suggesting that CoT provides useful intermediate reasoning signals that improve structural correctness. After introducing GRPO, the model achieves further improvements with a smoother, more stable trajectory, indicating that reward-based optimization can better correct sequence-level and global structural errors that are difficult to eliminate using token-level likelihood training alone.

\begin{table}[htbp]
	\centering
	\caption{NL-to-STL translation performance comparison.}
	\label{tab:nl2stl_compare}		
	\small 
	\renewcommand{\arraystretch}{1.2}
	
	\begin{tabularx}{\linewidth}{>{\raggedright\arraybackslash}X c c}
		\toprule
		\textbf{Method} & \textbf{Model size} & \textbf{Accuracy (\%)} \\
		\midrule
		Llama2-finetuned & 13B & 94.8 \\
		T5-finetuned & 220M & 93.1 \\
		Qwen2.5-0.5B (prompt-only) & 0.5B & 45.2 \\
		Qwen2.5-0.5B-finetuned w/o CoT & 0.5B & 93.3 \\
		\textbf{Qwen2.5-0.5B-finetuned w/ CoT (Proposed)} & \textbf{0.5B} & \textbf{98.0} \\
		\bottomrule
	\end{tabularx}
\end{table}

Quantitative results are summarized in Table~\ref{tab:nl2stl_compare}. The prompt-only Qwen2.5-0.5B-Instruct model achieves an accuracy of only $45.2\%$, highlighting the difficulty of reliably generating well-formed STL specifications without task-specific adaptation. Supervised fine-tuning without CoT already yields a substantial improvement, raising accuracy to $93.3\%$, which is comparable to or slightly better than the T5-finetuned baseline. The Llama2-finetuned model achieves $94.8\%$ accuracy but relies on a significantly larger parameter count.

By contrast, the proposed Qwen2.5-0.5B-Instruct model fine-tuned with CoT supervision and GRPO achieves an accuracy of $98.0\%$, outperforming all baselines while using substantially fewer parameters. Compared with the same backbone trained without CoT, this corresponds to an absolute improvement of $5.8$ percentage points, demonstrating that explicit reasoning supervision plays a critical role in reducing temporal-logic composition errors. These results indicate that for NL-to-STL translation, semantic alignment and structural reasoning are more decisive than raw model scale, and that the proposed reasoning-enhanced training pipeline is highly effective for producing executable formal specifications.

\subsection{Simulation Navigation Experiments}
\label{subsec:simulation_experiments} 
In this subsection, we evaluate the proposed NL navigation framework in simulation. All simulations are conducted in a $100,\mathrm{m}\times 100,\mathrm{m}$ planar workspace at a fixed altitude, with static obstacles and no-fly zones. The UAV dynamics follow the discrete-time linear model described in Section~II with sampling time $\Delta t=1,\mathrm{s}$, and the control limits enforce a maximum speed of $v_{\max}=5,\mathrm{m/s}$ and a maximum acceleration of $a_{\max}=1,\mathrm{m/s^2}$ throughout the planning horizon.
\subsubsection{Feasible STL Task Examples}
\label{subsec:feasible_tasks}

We first present representative task instances whose NL instructions can be translated into feasible STL specifications without any repair. These examples are used to illustrate the correspondence between language-level task descriptions, the translated STL formulas, and the resulting dynamically feasible trajectories.
\begin{figure}[h]
	\centering
	\includegraphics[width=0.5\textwidth,trim=0 5 0 5,	clip]{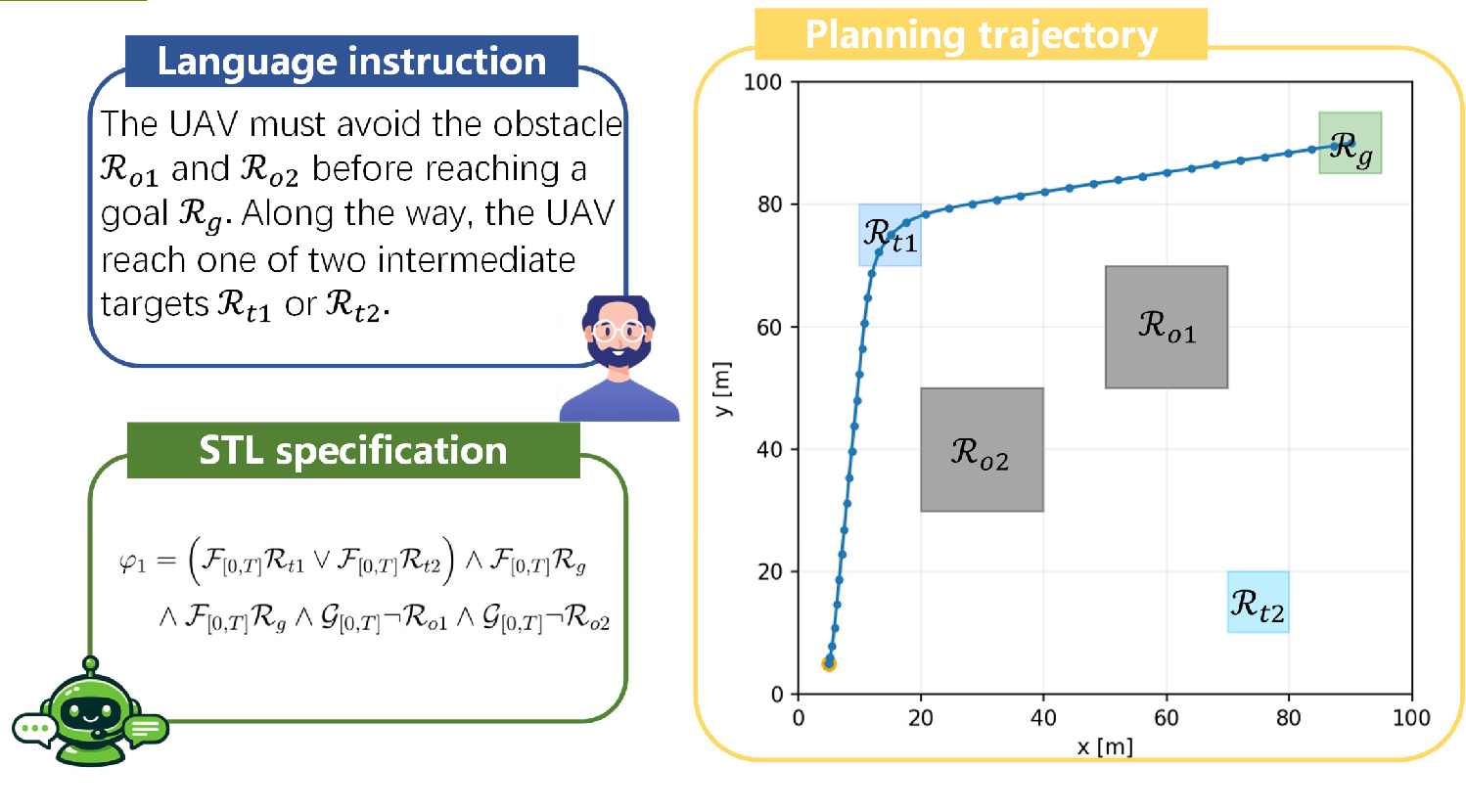}
	\caption{Simulated trajectory for Task~1. The UAV reaches the goal region within the specified time window while avoiding obstacles.}
	\label{fig:sim_task1}
\end{figure}

\textbf{Task 1 (Disjunctive intermediate goals with obstacle avoidance).}
The NL instruction is:
\begin{quote}
	``The UAV must avoid the obstacle $(\mathcal{R}_{o1})$ and $(\mathcal{R}_{o2})$ before reaching a goal $(\mathcal{R}_g)$. Along the way, the UAV must reach one of two intermediate targets $(\mathcal{R}_{t1})$ or $(\mathcal{R}_{t2})$.'' 
\end{quote}

This instruction is translated by the proposed translation module into the following STL specification:
\begin{align}
	\varphi_1 &=
	\Big(
	\mathcal{F}_{[0,T]}\mathcal{R}_{t1}
	\vee
	\mathcal{F}_{[0,T]}\mathcal{R}_{t2}
	\Big)\nonumber\\
	&\wedge
	\mathcal{F}_{[0,T]}\mathcal{R}_{g}
	\wedge
	\mathcal{G}_{[0,T]}\neg \mathcal{R}_{o1}
	\wedge
	\mathcal{G}_{[0,T]}\neg \mathcal{R}_{o2}.
\end{align}

The resulting STL specification is directly enforced by the MILP-based planner, which successfully finds a dynamically feasible trajectory satisfying all constraints. Fig.~\ref{fig:sim_task1} visualizes the planned trajectory, where the UAV reaches one of the intermediate targets and subsequently enters the goal region while maintaining safe clearance from the obstacle region.

\begin{figure}[h]
	\centering
	\includegraphics[width=0.5\textwidth,trim=3 5 0 5,	clip]{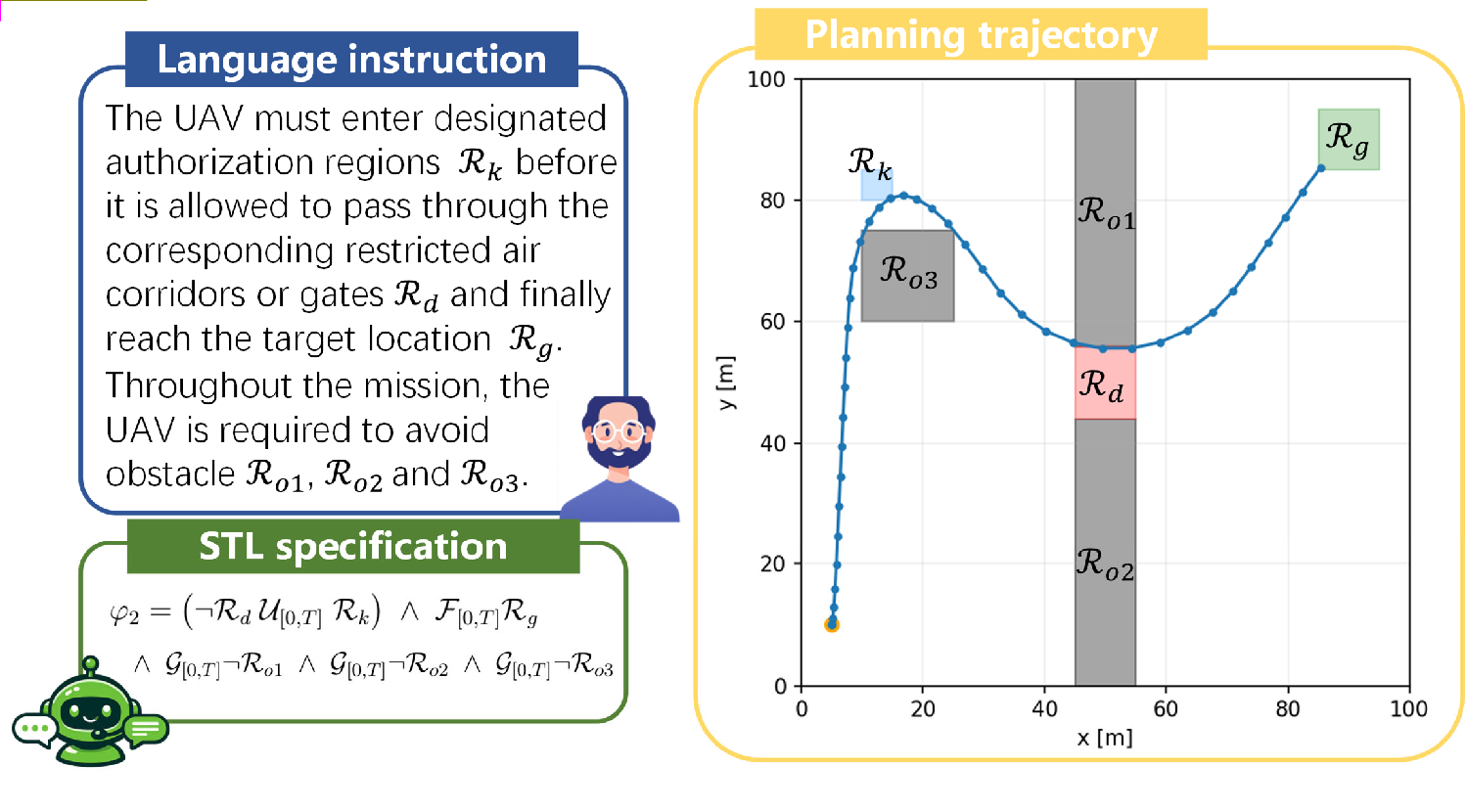}
	\caption{Simulated trajectory for Task~2. The UAV sequentially reaches the specified target regions while avoiding the no-fly zone at all times.}
	\label{fig:sim_task2}
\end{figure}

\textbf{Task 2 (Sequential authorization and constrained passage).}
The NL instruction is:
\begin{quote}
	``The UAV must enter designated authorization regions $\mathcal{R}_k$ before it is allowed to pass through the corresponding restricted air corridors or gates $\mathcal{R}_d$ and finally reach the target location $\mathcal{R}_g$. Throughout the mission, the UAV is required to avoid obstacle $(\mathcal{R}_{o1})$, $(\mathcal{R}_{o2})$ and $(\mathcal{R}_{o3})$.'' 
\end{quote}

This instruction is translated by the proposed translation module into the following STL specification:
\begin{align}
	\varphi_2 &=
	\big(
	\neg \mathcal{R}_{d}
	\;\mathcal{U}_{[0,T]}\;
	\mathcal{R}_{k}
	\big)
	\;\wedge\;
	\mathcal{F}_{[0,T]}\mathcal{R}_{g}\nonumber\\
	\;&\wedge\;
	\mathcal{G}_{[0,T]}\neg \mathcal{R}_{o1}
	\;\wedge\;
	\mathcal{G}_{[0,T]}\neg \mathcal{R}_{o2}
	\;\wedge\;
	\mathcal{G}_{[0,T]}\neg \mathcal{R}_{o3}.
\end{align}

Here, the until operator captures the temporal dependency that the UAV is prohibited from entering the restricted corridor or gate region $\mathcal{R}_d$ until it has visited the corresponding authorization region $\mathcal{R}_k$. Continuous avoidance of no-fly zones and obstacles is encoded as global safety constraints. The planner successfully synthesizes a trajectory that satisfies the sequential access requirement while respecting all safety constraints, as illustrated in Fig.~\ref{fig:sim_task2}.

\subsubsection{Specification Repair Example}
\label{subsec:repaired_task}

We next consider a representative UAV task in which a NL instruction induces an STL specification that is infeasible under the given dynamics and environmental constraints. This example demonstrates how the proposed language-guided specification repair mechanism restores feasibility by selectively relaxing temporal requirements, while preserving all safety-critical constraints and geometric task definitions.

\begin{figure}[h]
	\centering
	\includegraphics[width=0.5\textwidth,trim=0 5 0 5,clip]{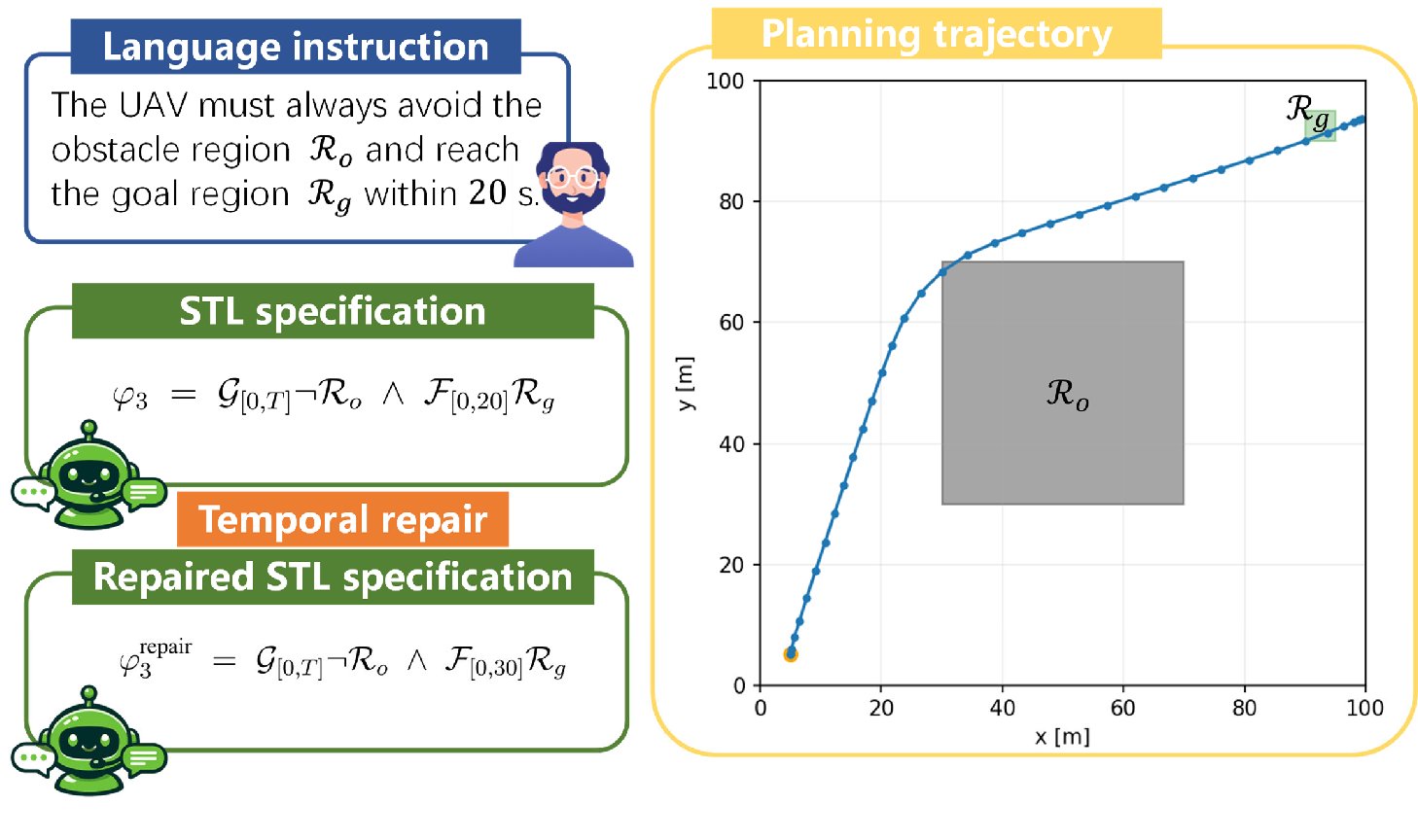}
	\caption{Simulated trajectory for the repaired task. The UAV reaches the goal region after temporal repair while continuously avoiding the obstacle.}
	\label{fig:sim_task3_repaired}
\end{figure}

\begin{figure*}[t]
	\centering
	\includegraphics[width=0.8\textwidth,trim=0 0 0 0,clip]{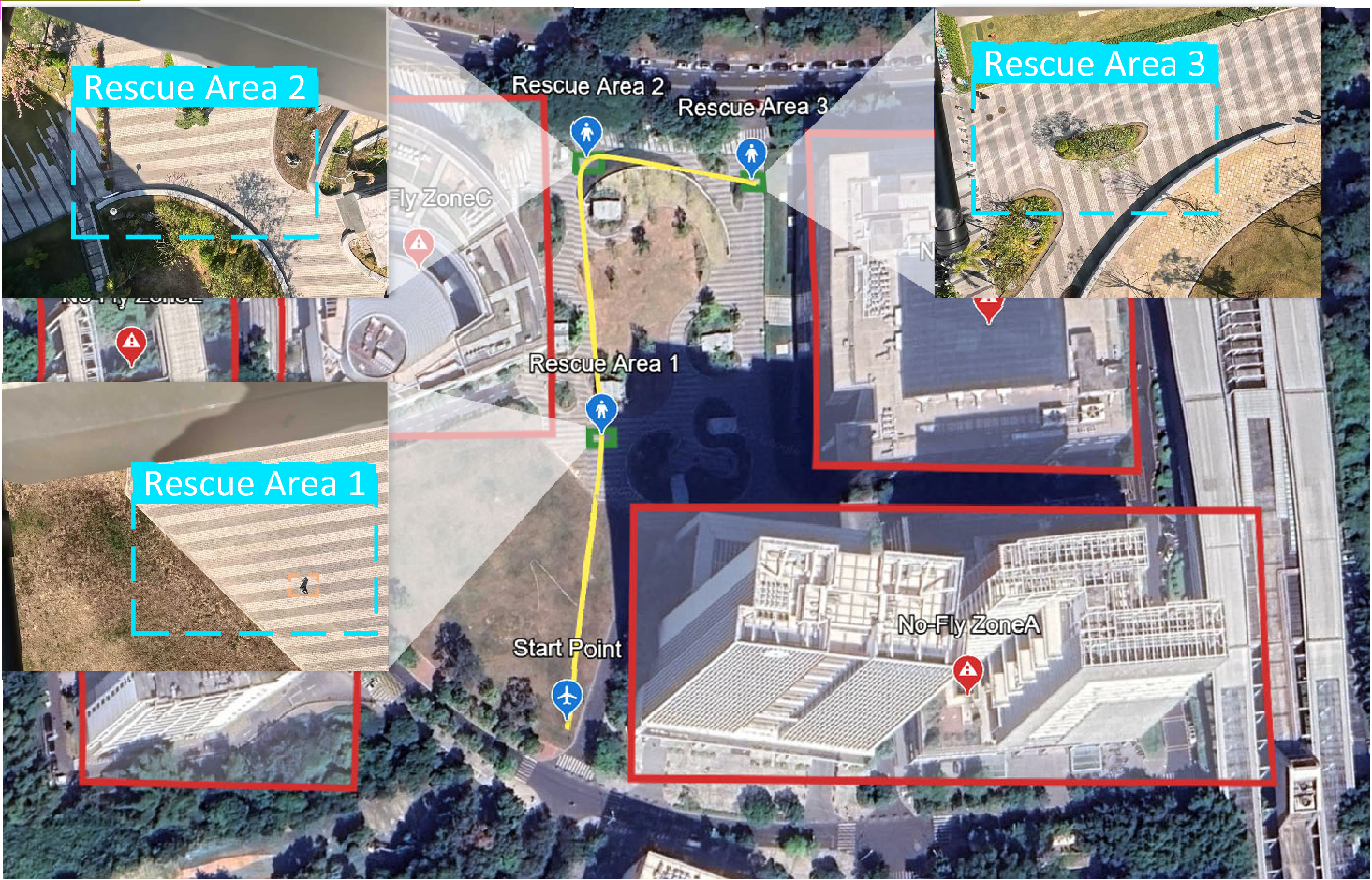}
	\caption{Executed UAV trajectory for the real-world search-and-rescue experiment.}
	\label{fig:realworld_trajectory}
\end{figure*}

\textbf{Task 3 (Time-constrained goal reaching with obstacle avoidance).}
The NL instruction provided by the human operator is:
\begin{quote}
	``The UAV must always avoid the obstacle region $(\mathcal{R}_{o})$ and reach the goal region $(\mathcal{R}_{g})$ within $20\,\mathrm{s}$.'' 
\end{quote}

This instruction is translated by the proposed translation module into the following STL specification:
\begin{equation}
	\varphi_3 \;=\;
	\mathcal{G}_{[0,T]}\neg \mathcal{R}_{o}
	\;\wedge\;
	\mathcal{F}_{[0,20]}\mathcal{R}_{g},
\end{equation}
where $\mathcal{R}_{o}$ denotes the obstacle region and $\mathcal{R}_{g}$ denotes the goal region. The global operator $\mathcal{G}$ enforces continuous obstacle avoidance over the entire planning horizon, while the eventually operator $\mathcal{F}_{[0,20]}$ requires the UAV to reach the goal within a strict deadline of $20\,\mathrm{s}$.

Under the system dynamics and control limits described earlier in this section, the resulting MILP problem is infeasible. The obstacle lies directly between the initial position and the goal, forcing the UAV to execute a detour. Given the bounded velocity and acceleration, the solver determines that the UAV cannot reach $\mathcal{R}_{g}$ within the specified $20\,\mathrm{s}$ without violating either the dynamic constraints or the obstacle-avoidance requirement.

Upon detecting infeasibility, the specification repair procedure is automatically triggered. The MILP solver computes an  IIS, and the conflicting constraints are traced back to the STL subformulas using the constraint-to-specification mapping. In this case, the conflict is localized to the temporal bound of the reachability requirement $\mathcal{F}_{[0,20]}\mathcal{R}_{g}$.

Guided by the LLM, the repair module is instructed to relax \emph{only} the temporal component of the task, while keeping all geometric predicates unchanged. In particular, the size and location of the goal region $\mathcal{R}_{g}$ and the obstacle-avoidance constraint $\mathcal{G}_{[0,T]}\neg \mathcal{R}_{o}$ are treated as safety-critical and are not modified. A slack variable is introduced on the upper bound of the eventually operator, and a weighted $\ell_1$ penalty is used to minimize the amount of temporal relaxation.

As a result, the repaired specification becomes:
\begin{equation}
	\varphi_3^{\text{repair}} \;=\;
	\mathcal{G}_{[0,T]}\neg \mathcal{R}_{o}
	\;\wedge\;
	\mathcal{F}_{[0,30]}\mathcal{R}_{g},
\end{equation}
corresponding to an extension of the reachability deadline from $20\,\mathrm{s}$ to $30\,\mathrm{s}$.

The repaired STL specification is then enforced by the MILP-based planner, which successfully synthesizes a dynamically feasible trajectory. As shown in Fig.~\ref{fig:sim_task3_repaired}, the UAV detours around $\mathcal{R}_{o}$, maintains continuous safety, and reaches the original goal region $\mathcal{R}_{g}$ within the relaxed temporal window. This example illustrates that the proposed framework can reconcile high-level time-critical language instructions with low-level dynamical feasibility through targeted, interpretable specification repair, without compromising safety constraints or task semantics.

\begin{figure}[h]
	\centering
	\includegraphics[width=0.5\textwidth]{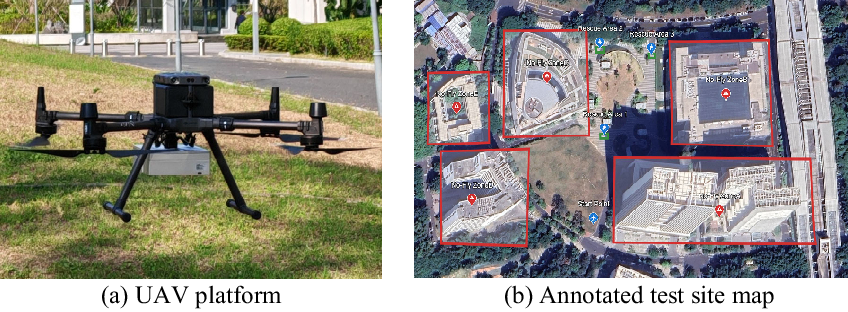}
	\caption{Real-world experimental setup for the search-and-rescue task. (a) DJI Matrice~300 RTK equipped with an onboard 4G/5G wireless signal acquisition and localization device. (b) Annotated test site map showing designated rescue areas and no-fly zones.}
	\label{fig:realworld_setup}
\end{figure}

\subsection{Real-World Experiments}
\label{subsec:realworld_experiments}

To validate the proposed framework under real-world conditions, we conduct a real-World experiment involving an UAV performing a search-and-rescue task guided by NL instructions.

\subsubsection{Experimental Setup}
\label{subsec:realworld_setup}

The experiments are carried out using a DJI Matrice~300 RTK quadrotor equipped with an onboard 4G/5G wireless signal acquisition and localization device. The UAV operates in an outdoor environment with a predefined bounded workspace that includes designated search-and-rescue regions and safety-restricted no-fly zones. State estimation is provided by the onboard RTK positioning system and fused inertial measurements, while LLM translation and STL planning are performed offboard and transmitted to the UAV via a wireless communication link. Fig.~\ref{fig:realworld_setup} illustrates the real-world experimental setup, including the UAV platform and the annotated test site with rescue areas and no-fly zones.

\subsubsection{Search-and-Rescue Task Description}
\label{subsec:realworld_task}

We consider a representative search-and-rescue scenario in which the UAV is instructed to search multiple designated regions within a fixed time window while continuously avoiding all no-fly zones. The task is specified through the following NL instruction:
\begin{itemize}
	\item ``Search the three rescue areas $\mathcal{R}_{s1}$, $\mathcal{R}_{s2}$, and $\mathcal{R}_{s3}$ within 60 seconds, while avoiding all no-fly zones.''
\end{itemize}

This instruction is translated by the proposed translation module into the following STL specification:
\begin{align}
	\label{eq:phi_sar}
	\varphi_{\mathrm{SAR}} &=
	\mathcal{F}_{[0,60]}\mathcal{R}_{s1}
	\;\wedge\;
	\mathcal{F}_{[0,60]}\mathcal{R}_{s2}
	\;\wedge\;
	\mathcal{F}_{[0,60]}\mathcal{R}_{s3} \nonumber\\
	&\quad \;\wedge\;
	\mathcal{G}_{[0,60]} \neg \mathcal{R}_{nf1}
	\;\wedge\;
	\mathcal{G}_{[0,60]} \neg \mathcal{R}_{nf2}
	\;\wedge\;
	\mathcal{G}_{[0,60]} \neg \mathcal{R}_{nf3} \nonumber\\
	&\quad \;\wedge\;
	\mathcal{G}_{[0,60]} \neg \mathcal{R}_{nf4}
	\;\wedge\;
	\mathcal{G}_{[0,60]} \neg \mathcal{R}_{nf5}.
\end{align}

Here, $\mathcal{R}_{s1}$, $\mathcal{R}_{s2}$, and $\mathcal{R}_{s3}$ denote the three designated search regions, while $\mathcal{R}_{nf1}$--$\mathcal{R}_{nf5}$ represent the individual no-fly zones in the environment. The global operator $\mathcal{G}_{[0,60]}$ enforces continuous avoidance of all no-fly regions over the entire mission horizon, encoding safety-critical constraints that must never be violated. The eventually operators $\mathcal{F}_{[0,60]}$ require that each search region be visited at least once within the 60-second time window, without imposing a strict visitation order among them.

The resulting STL specification captures the essential requirements of the search-and-rescue task, namely time-bounded coverage of all designated search areas and persistent avoidance of restricted airspace, and is subsequently enforced by the MILP-based planner to generate a dynamically feasible patrol trajectory.

\subsubsection{Experimental Results}
\label{subsec:realworld_results}

Fig.~\ref{fig:realworld_trajectory} shows the executed patrol trajectory during the search-and-rescue task. The UAV successfully completes the patrol within the prescribed time window while respecting all safety constraints specified by the STL formula. During the mission, the UAV detected a victim waiting for rescue inside the region $\mathcal{R}_{s1}$ and reported the finding to the ground operator for subsequent response. The experiment demonstrates that the proposed language-guided planning framework can be deployed on a real UAV system and can reliably execute complex, temporally constrained tasks derived from NL instructions.

\section{Conclusion}
\label{sec:conclusion}

This paper presented a unified framework for NL low-altitude UAV navigation by translating free-form instructions into STL specifications and synthesizing dynamically feasible trajectories under formal constraints. By integrating a reasoning-enhanced LLM with MILP-based STL-constrained planning, the proposed approach enables robust NL-to-STL translation while rigorously enforcing safety-critical requirements. A solver-in-the-loop specification repair mechanism was further introduced, in which an LLM provides semantic guidance to selectively relax non-safety-critical task constraints while strictly preserving safety guarantees. Extensive simulation results and real-world flight experiments demonstrate that the proposed framework achieves safe, interpretable, and adaptable UAV navigation in complex low-altitude environments. Future work will focus on extending the framework to partially observable environments, multi-UAV coordination, and online adaptation under dynamic task revisions.

\bibliographystyle{IEEEtran}
\bibliography{ref}

@article{hari2020optimal,
  title={Optimal {UAV} route planning for persistent monitoring missions},
  author={Hari, Sai Krishna Kanth and Rathinam, Sivakumar and Darbha, Swaroop and Kalyanam, Krishna and Manyam, Satyanarayana Gupta and Casbeer, David},
  journal={IEEE Transactions on Robotics},
  volume={37},
  number={2},
  pages={550--566},
  year={2020},
  publisher={IEEE}
}

@ARTICLE{pyq2025mllmuav,
  author={Ping, Yuqi and Liang, Tianhao and Ding, Huahao and Lei, Guangyu and Wu, Junwei and Zou, Xuan and Shi, Kuan and Shao, Rui and Zhang, Chiya and Zhang, Weizheng and Yuan, Weijie and Zhang, Tingting},
  journal={IEEE Wireless Communications}, 
  title={Multimodal Large Language Models-Enabled {UAV} Swarm: Towards Efficient and Intelligent Autonomous Aerial Systems}, 
  year={2025},
  volume={},
  number={},
  pages={1-9},
  doi={10.1109/MWC.2025.3630121}}

@ARTICLE{11313542,
  author={Chen, Mingjian and Yang, Liang and Cao, Jiangling and Zhu, Guangxu and Yuan, Weijie and Jiang, Hongbo and Niyato, Dusit},
  journal={IEEE Transactions on Mobile Computing}, 
  title={Cargo {UAVs} Pick-up Systems for Low-Altitude Economy With Communication Quality, Battery Energy, and Time Window Constraints}, 
  year={2025},
  volume={},
  number={},
  pages={1-18},
  doi={10.1109/TMC.2025.3647000}}

@article{liang2025satellite,
  title={Satellite-Assisted {UAV} Control: Sensing and Communication Scheduling for Energy Efficient Data Collection},
  author={Liang, Tianhao and Ding, Huahao and Ping, Yuqi and Zhang, Tingting and Zhou, Longyu and Zhang, Qinyu and Quek, Tony QS},
  journal={IEEE Internet of Things Journal},
  year={2025},
  publisher={IEEE}
}

@article{liu2023tangent,
  title={Tangent-based path planning for {UAV} in a {3-D} low altitude urban environment},
  author={Liu, Huan and Wu, Guohua and Zhou, Ling and Pedrycz, Witold and Suganthan, Ponnuthurai Nagaratnam},
  journal={IEEE Transactions on Intelligent Transportation Systems},
  volume={24},
  number={11},
  pages={12062--12077},
  year={2023},
  publisher={IEEE}
}

@inproceedings{quartey2025verifiably,
  title={Verifiably following complex robot instructions with foundation models},
  author={Quartey, Benedict and Rosen, Eric and Tellex, Stefanie and Konidaris, George},
  booktitle={2025 IEEE International Conference on Robotics and Automation (ICRA)},
  pages={1--8},
  year={2025},
  organization={IEEE}
}

@article{tellex2020robots,
  title={Robots that use language},
  author={Tellex, Stefanie and Gopalan, Nakul and Kress-Gazit, Hadas and Matuszek, Cynthia},
  journal={Annual Review of Control, Robotics, and Autonomous Systems},
  volume={3},
  number={1},
  pages={25--55},
  year={2020},
  publisher={Annual Reviews}
}

@article{belta2019formal,
  title={Formal methods for control synthesis: An optimization perspective},
  author={Belta, Calin and Sadraddini, Sadra},
  journal={Annual Review of Control, Robotics, and Autonomous Systems},
  volume={2},
  number={1},
  pages={115--140},
  year={2019},
  publisher={Annual Reviews}
}

@inproceedings{liu2023aerialvln,
  title={{AerialVLN}: Vision-and-language navigation for {UAVs}},
  author={Liu, Shubo and Zhang, Hongsheng and Qi, Yuankai and Wang, Peng and Zhang, Yanning and Wu, Qi},
  booktitle={Proceedings of the IEEE/CVF International Conference on Computer Vision},
  pages={15384--15394},
  year={2023}
}

@inproceedings{chandarana2017natural,
  title={Natural language based multimodal interface for {UAV} mission planning},
  author={Chandarana, Meghan and Meszaros, Erica L and Trujillo, Anna and Danette Allen, B},
  booktitle={Proceedings of the Human Factors and Ergonomics Society Annual Meeting},
  volume={61},
  number={1},
  pages={68--72},
  year={2017},
  organization={SAGE Publications Sage CA: Los Angeles, CA}
}

@inproceedings{chandarana2017fly,
  title={{'Fly Like This'}: Natural Language Interface for UAV Mission Planning},
  author={Chandarana, Meghan and Meszaros, Erica L and Trujillo, Anna and Allen, B Danette},
  booktitle={International Conference on Advances in Computer-Human Interactions},
  number={NF1676L-26108},
  year={2017}
}

@article{yao2025aeroverse,
  title={{AeroVerse-Review}: Comprehensive survey on aerial embodied vision-and-language navigation},
  author={Yao, Fanglong and Liu, Youzhi and Zhang, Wenyi and Zhu, Zhengqiu and Li, Chenglong and Liu, Nayu and Hu, Peng and Yue, Yuanchang and Wei, Kaiwen and He, Xin and others},
  journal={The Innovation Informatics},
  volume={1},
  number={1},
  pages={100015--1},
  year={2025},
  publisher={The Innovation Informatics}
}

@inproceedings{anderson2018vision,
  title={Vision-and-language navigation: Interpreting visually-grounded navigation instructions in real environments},
  author={Anderson, Peter and Wu, Qi and Teney, Damien and Bruce, Jake and Johnson, Mark and S{\"u}nderhauf, Niko and Reid, Ian and Gould, Stephen and Van Den Hengel, Anton},
  booktitle={Proceedings of the IEEE conference on computer vision and pattern recognition},
  pages={3674--3683},
  year={2018}
}

@inproceedings{narasimhan2015language,
    title = "Language Understanding for Text-based Games using Deep Reinforcement Learning",
    author = "Narasimhan, Karthik  and
      Kulkarni, Tejas  and
      Barzilay, Regina",
    editor = "M{\`a}rquez, Llu{\'i}s  and
      Callison-Burch, Chris  and
      Su, Jian",
    booktitle = "Proceedings of the 2015 Conference on Empirical Methods in Natural Language Processing",
    month = sep,
    year = "2015",
    address = "Lisbon, Portugal",
    publisher = "Association for Computational Linguistics",
    url = "https://aclanthology.org/D15-1001/",
    doi = "10.18653/v1/D15-1001",
    pages = "1--11"
}

@inproceedings{wang2019reinforced,
  title={Reinforced cross-modal matching and self-supervised imitation learning for vision-language navigation},
  author={Wang, Xin and Huang, Qiuyuan and Celikyilmaz, Asli and Gao, Jianfeng and Shen, Dinghan and Wang, Yuan-Fang and Wang, William Yang and Zhang, Lei},
  booktitle={Proceedings of the IEEE/CVF conference on computer vision and pattern recognition},
  pages={6629--6638},
  year={2019}
}

@inproceedings{hong2021vln,
  title={{VLN BERT}: A recurrent vision-and-language {BERT} for navigation},
  author={Hong, Yicong and Wu, Qi and Qi, Yuankai and Rodriguez-Opazo, Cristian and Gould, Stephen},
  booktitle={Proceedings of the IEEE/CVF conference on Computer Vision and Pattern Recognition},
  pages={1643--1653},
  year={2021}
}

@article{zhang2025citynavagent,
  title={{CityNavAgent}: Aerial Vision-and-Language Navigation with Hierarchical Semantic Planning and Global Memory},
  author={Zhang, Weichen and Gao, Chen and Yu, Shiquan and Peng, Ruiying and Zhao, Baining and Zhang, Qian and Cui, Jinqiang and Chen, Xinlei and Li, Yong},
  journal={arXiv preprint arXiv:2505.05622},
  year={2025}
}

@article{saxena2025uav,
  title={{UAV-VLN}: {End-to-End} Vision Language guided Navigation for UAVs},
  author={Saxena, Pranav and Raghuvanshi, Nishant and Goveas, Neena},
  journal={arXiv preprint arXiv:2504.21432},
  year={2025}
}

@article{lee2024citynav,
  title={Citynav: Language-goal aerial navigation dataset with geographic information},
  author={Lee, Jungdae and Miyanishi, Taiki and Kurita, Shuhei and Sakamoto, Koya and Azuma, Daichi and Matsuo, Yutaka and Inoue, Nakamasa},
  journal={arXiv preprint arXiv:2406.14240},
  year={2024}
}

@article{sanyal2025asma,
  title={Asma: An adaptive safety margin algorithm for vision-language drone navigation via scene-aware control barrier functions},
  author={Sanyal, Sourav and Roy, Kaushik},
  journal={IEEE Robotics and Automation Letters},
  year={2025},
  publisher={IEEE}
}

@inproceedings{wu2025selp,
  title={{SELP}: Generating safe and efficient task plans for robot agents with large language models},
  author={Wu, Yi and Xiong, Zikang and Hu, Yiran and Iyengar, Shreyash S and Jiang, Nan and Bera, Aniket and Tan, Lin and Jagannathan, Suresh},
  booktitle={2025 IEEE International Conference on Robotics and Automation (ICRA)},
  pages={2599--2605},
  year={2025},
  organization={IEEE}
}

@article{xu2025nl2hltl2plan,
  title={Nl2hltl2plan: Scaling up natural language understanding for multi-robots through hierarchical temporal logic task specifications},
  author={Xu, Shaojun and Luo, Xusheng and Huang, Yutong and Leng, Letian and Liu, Ruixuan and Liu, Changliu},
  journal={IEEE Robotics and Automation Letters},
  year={2025},
  publisher={IEEE}
}

@INPROCEEDINGS{10610123,
  author={Zhang, Yixiao and Fernandez-Ayala, Victor Nan and Dimarogonas, Dimos V.},
  booktitle={2024 IEEE International Conference on Robotics and Automation (ICRA)}, 
  title={Multi-robot Human-in-the-loop Control under Spatiotemporal Specifications}, 
  year={2024},
  volume={},
  number={},
  pages={4841-4847},
  doi={10.1109/ICRA57147.2024.10610123}}

@inproceedings{kress2007structured,
  title={From structured english to robot motion},
  author={Kress-Gazit, Hadas and Fainekos, Georgios E and Pappas, George J},
  booktitle={2007 IEEE/RSJ International Conference on Intelligent Robots and Systems},
  pages={2717--2722},
  year={2007},
  organization={IEEE}
}

@inproceedings{patel2020grounding,
  title={Grounding Language to {Non-Markovian} Tasks with No Supervision of Task Specifications.},
  author={Patel, Roma and Pavlick, Ellie and Tellex, Stefanie},
  booktitle={Robotics: Science and Systems},
  volume={2020},
  year={2020}
}

@inproceedings{wang2021learning,
  title={Learning a natural-language to {LTL} executable semantic parser for grounded robotics},
  author={Wang, Christopher and Ross, Candace and Kuo, Yen-Ling and Katz, Boris and Barbu, Andrei},
  booktitle={Conference on Robot Learning},
  pages={1706--1718},
  year={2021},
  organization={PMLR}
}

@inproceedings{gopalan2018sequence,
  title={{Sequence-to-Sequence} Language Grounding of {Non-Markovian} Task Specifications.},
  author={Gopalan, Nakul and Arumugam, Dilip and Wong, Lawson LS and Tellex, Stefanie},
  booktitle={Robotics: Science and Systems},
  volume={2018},
  year={2018}
}

@article{pan2023data,
  title={Data-efficient learning of natural language to linear temporal logic translators for robot task specification},
  author={Pan, Jiayi and Chou, Glen and Berenson, Dmitry},
  journal={arXiv preprint arXiv:2303.08006},
  year={2023}
}

@inproceedings{fuggitti2023nl2ltl,
  title={Nl2ltl--a python package for converting natural language instructions to linear temporal logic  formulas},
  author={Fuggitti, Francesco and Chakraborti, Tathagata},
  booktitle={Proceedings of the AAAI Conference on Artificial Intelligence},
  volume={37},
  number={13},
  pages={16428--16430},
  year={2023}
}

@inproceedings{liu2023grounding,
  title={Grounding complex natural language commands for temporal tasks in unseen environments},
  author={Liu, Jason Xinyu and Yang, Ziyi and Idrees, Ifrah and Liang, Sam and Schornstein, Benjamin and Tellex, Stefanie and Shah, Ankit},
  booktitle={Conference on Robot Learning},
  pages={1084--1110},
  year={2023},
  organization={PMLR}
}

@inproceedings{ghosh2016diagnosis,
  title={Diagnosis and repair for synthesis from signal temporal logic specifications},
  author={Ghosh, Shromona and Sadigh, Dorsa and Nuzzo, Pierluigi and Raman, Vasumathi and Donz{\'e}, Alexandre and Sangiovanni-Vincentelli, Alberto L and Sastry, S Shankar and Seshia, Sanjit A},
  booktitle={Proceedings of the 19th International Conference on Hybrid Systems: Computation and Control},
  pages={31--40},
  year={2016}
}

@inproceedings{buyukkocak2022temporal,
  title={Temporal relaxation of signal temporal logic specifications for resilient control synthesis},
  author={Buyukkocak, Ali Tevfik and Aksaray, Derya},
  booktitle={2022 IEEE 61st Conference on Decision and Control (CDC)},
  pages={2890--2896},
  year={2022},
  organization={IEEE}
}

@article{buyukkocak2025resilient,
  title={Resilient Online Planning for Mobile Robots with Minimal Relaxation of Signal Temporal Logic Specifications},
  author={Buyukkocak, Ali Tevfik and Aksaray, Derya},
  journal={IEEE Robotics and Automation Letters},
  year={2025},
  publisher={IEEE}
}

@inproceedings{donze2010robust,
  title={Robust satisfaction of temporal logic over real-valued signals},
  author={Donz{\'e}, Alexandre and Maler, Oded},
  booktitle={International conference on formal modeling and analysis of timed systems},
  pages={92--106},
  year={2010},
  organization={Springer}
}

@misc{deepseekmath,
      title={DeepSeek-V3 Technical Report}, 
      author = {{DeepSeek-AI} and Aixin Liu and Bei Feng and Bing Xue and Bingxuan Wang and Bochao Wu and {et al.}},
      eprint={2412.19437},
      archivePrefix={arXiv},
      primaryClass={cs.CL},
      url={https://arxiv.org/abs/2412.19437}, 
}

@misc{deepseekr1,
      title={{DeepSeek-R1}: Incentivizing Reasoning Capability in {LLMs} via Reinforcement Learning}, 
      author={{DeepSeek-AI} and Daya Guo and Dejian Yang and Haowei Zhang and Junxiao Song and Ruoyu Zhang and and {et al.}},
      year={2025},
      eprint={2501.12948},
      archivePrefix={arXiv},
      primaryClass={cs.CL},
      url={https://arxiv.org/abs/2501.12948}, 
}

@misc{PPO,
      title={Proximal Policy Optimization Algorithms}, 
      author={John Schulman and Filip Wolski and Prafulla Dhariwal and Alec Radford and Oleg Klimov},
      year={2017},
      eprint={1707.06347},
      archivePrefix={arXiv},
      primaryClass={cs.LG},
      url={https://arxiv.org/abs/1707.06347}, 
}

@article{zhang2025toward,
  title={Toward edge general intelligence with agentic {AI} and agentification: Concepts, technologies, and future directions},
  author={Zhang, Ruichen and Liu, Guangyuan and Liu, Yinqiu and Zhao, Changyuan and Wang, Jiacheng and Xu, Yunting and Niyato, Dusit and Kang, Jiawen and Li, Yonghui and Mao, Shiwen and others},
  journal={arXiv preprint arXiv:2508.18725},
  year={2025}
}

@article{zhang2024interactive,
  title={Interactive {AI} with retrieval-augmented generation for next generation networking},
  author={Zhang, Ruichen and Du, Hongyang and Liu, Yinqiu and Niyato, Dusit and Kang, Jiawen and Sun, Sumei and Shen, Xuemin and Poor, H Vincent},
  journal={IEEE Network},
  volume={38},
  number={6},
  pages={414--424},
  year={2024},
  publisher={IEEE}
}

@ARTICLE{zhang2025covert,
  author={Zhang, Ruichen and Liu, Yinqiu and Tang, Shunpu and Wang, Jiacheng and Niyato, Dusit and Sun, Geng and Li, Yonghui and Sun, Sumei},
  journal={IEEE Journal on Selected Areas in Communications}, 
  title={Covert Prompt Transmission for Secure Large Language Model Services}, 
  year={2025},
  volume={},
  number={},
  pages={1-1},
  doi={10.1109/JSAC.2025.3647412}}

@ARTICLE{11303308,
  author={Zhang, Ruichen and Tang, Shunpu and Liu, Yinqiu and Niyato, Dusit and Xiong, Zehui and Sun, Sumei and Mao, Shiwen and Han, Zhu},
  journal={IEEE Communications Magazine}, 
  title={Toward Agentic {AI}: Generative Information Retrieval Inspired Intelligent Communications and Networking}, 
  year={2026},
  volume={64},
  number={1},
  pages={197-204},
  doi={10.1109/MCOM.001.2500073}}

@article{zhang2024generative,
  title={Generative {AI} agents with large language model for satellite networks via a mixture of experts transmission},
  author={Zhang, Ruichen and Du, Hongyang and Liu, Yinqiu and Niyato, Dusit and Kang, Jiawen and Xiong, Zehui and Jamalipour, Abbas and Kim, Dong In},
  journal={IEEE Journal on Selected Areas in Communications},
  year={2024},
  publisher={IEEE}
}

@article{shao2025large,
  title={Large {VLM}-based vision-language-action models for robotic manipulation: A survey},
  author={Shao, Rui and Li, Wei and Zhang, Lingsen and Zhang, Renshan and Liu, Zhiyang and Chen, Ran and Nie, Liqiang},
  journal={arXiv preprint arXiv:2508.13073},
  year={2025}
}

@article{chen2023nl2tl,
  title={Nl2tl: Transforming natural languages to temporal logics using large language models},
  author={Chen, Yongchao and Gandhi, Rujul and Zhang, Yang and Fan, Chuchu},
  journal={arXiv preprint arXiv:2305.07766},
  year={2023}
}

@article{hu2022lora,
  title={{LORA}: Low-rank adaptation of large language models.},
  author={Hu, Edward J and Shen, Yelong and Wallis, Phillip and Allen-Zhu, Zeyuan and Li, Yuanzhi and Wang, Shean and Wang, Lu and Chen, Weizhu and others},
  journal={ICLR},
  volume={1},
  number={2},
  pages={3},
  year={2022}
}

@inproceedings{papineni2002bleu,
  title={{BLEU}: a method for automatic evaluation of machine translation},
  author={Papineni, Kishore and Roukos, Salim and Ward, Todd and Zhu, Wei-Jing},
  booktitle={Proceedings of the 40th annual meeting of the Association for Computational Linguistics},
  pages={311--318},
  year={2002}
}

@inproceedings{mao2024nl2stl,
  title={Nl2stl: Transformation from logic natural language to signal temporal logics using llama2},
  author={Mao, Yuchen and Zhang, Tianci and Cao, Xu and Chen, Zhongyao and Liang, Xinkai and Xu, Bochen and Fang, Hao},
  booktitle={2024 IEEE International Conference on Cybernetics and Intelligent Systems (CIS) and IEEE International Conference on Robotics, Automation and Mechatronics (RAM)},
  pages={469--474},
  year={2024},
  organization={IEEE}
}

\end{document}